\PassOptionsToPackage{table}{xcolor}
\documentclass{article}

\usepackage{microtype}
\usepackage{graphicx}
\usepackage{subfigure}
\usepackage{booktabs} 
\usepackage{multirow}
\usepackage{makecell}

\usepackage{amsmath, amssymb, amscd, amsthm, amsfonts}
\usepackage{bm}
\usepackage{graphicx} 

\usepackage{hyperref}

\usepackage[accepted]{icml2025}

\usepackage{amsmath}
\usepackage{amssymb}
\usepackage{mathtools}
\usepackage{amsthm}
\usepackage{algorithm}
\usepackage{algorithmic}

\usepackage{amsmath,epstopdf,amssymb,color,url,multirow,multicol,verbatim,threeparttable,diagbox,makecell,booktabs}
\definecolor{light-gray}{gray}{0.82}
\definecolor{aliceblue}{rgb}{0.94,0.97,1.0}

\usepackage[capitalize,noabbrev]{cleveref}

\theoremstyle{plain}

\theoremstyle{definition}

\theoremstyle{remark}

\usepackage[textsize=tiny]{todonotes}

\icmltitlerunning{Orthogonal Subspace Decomposition for Generalizable AI-Generated Image Detection}

\begin{document}

\twocolumn[
\icmltitle{Orthogonal Subspace Decomposition for Generalizable AI-Generated Image Detection}



\icmlsetsymbol{equal}{*}

\begin{icmlauthorlist}
\icmlauthor{Zhiyuan Yan}{equal,yyy,comp}
\icmlauthor{Jiangming Wang}{equal,comp}
\icmlauthor{Peng Jin}{yyy}
\icmlauthor{Ke-Yue Zhang}{comp}
\icmlauthor{Chengchun Liu}{yyy}
\icmlauthor{Shen Chen}{comp}
\icmlauthor{Taiping Yao}{comp}
\icmlauthor{Shouhong Ding}{comp}
\icmlauthor{Baoyuan Wu}{sch}
\icmlauthor{Li Yuan}{yyy}
\end{icmlauthorlist}


\icmlaffiliation{yyy}{Peking University Shenzhen Graduate School}
\icmlaffiliation{comp}{Tencent Youtu Lab}
\icmlaffiliation{sch}{The Chinese University of Hong Kong, Shenzhen}

\icmlcorrespondingauthor{Taiping Yao}{taipingyao@tencent.cn}
\icmlcorrespondingauthor{Li Yuan}{yuanli-ece@pku.edu.cn}


\icmlkeywords{AI-generated Image Detection, Image Decomposition, low-level vision, Deep Learning}

\vskip 0.3in
]



\printAffiliationsAndNotice{\icmlEqualContribution} 

\begin{abstract}
AI-generated images (AIGIs), such as natural or face images, have become increasingly important yet challenging.
In this paper, we start from a new perspective to excavate the reason behind the failure generalization in AIGI detection, named the \textit{asymmetry phenomenon}, where a naively trained detector tends to favor overfitting to the limited and monotonous fake patterns, causing the feature space to become highly constrained and low-ranked, which is proved seriously limiting the expressivity and generalization.
One potential remedy is incorporating the pre-trained knowledge within the vision foundation models (higher-ranked) to expand the feature space, alleviating the model's overfitting to fake.
To this end, we employ Singular Value Decomposition (SVD) to decompose the original feature space into \textit{two orthogonal subspaces}.
By freezing the principal components and adapting only the remained components, we preserve the pre-trained knowledge while learning fake patterns.
Compared to existing full-parameters and LoRA-based tuning methods, we explicitly ensure orthogonality, enabling the higher rank of the whole feature space, effectively minimizing overfitting and enhancing generalization.
We finally identify a crucial insight: our method implicitly learns \textit{a vital prior that fakes are actually derived from the real}, indicating a hierarchical relationship rather than independence. Modeling this prior, we believe, is essential for achieving superior generalization.
Our codes are publicly available at \href{https://github.com/YZY-stack/Effort-AIGI-Detection}{GitHub}.
\end{abstract}

\begin{figure}[t]
    \centering
    \includegraphics[width=1\linewidth]{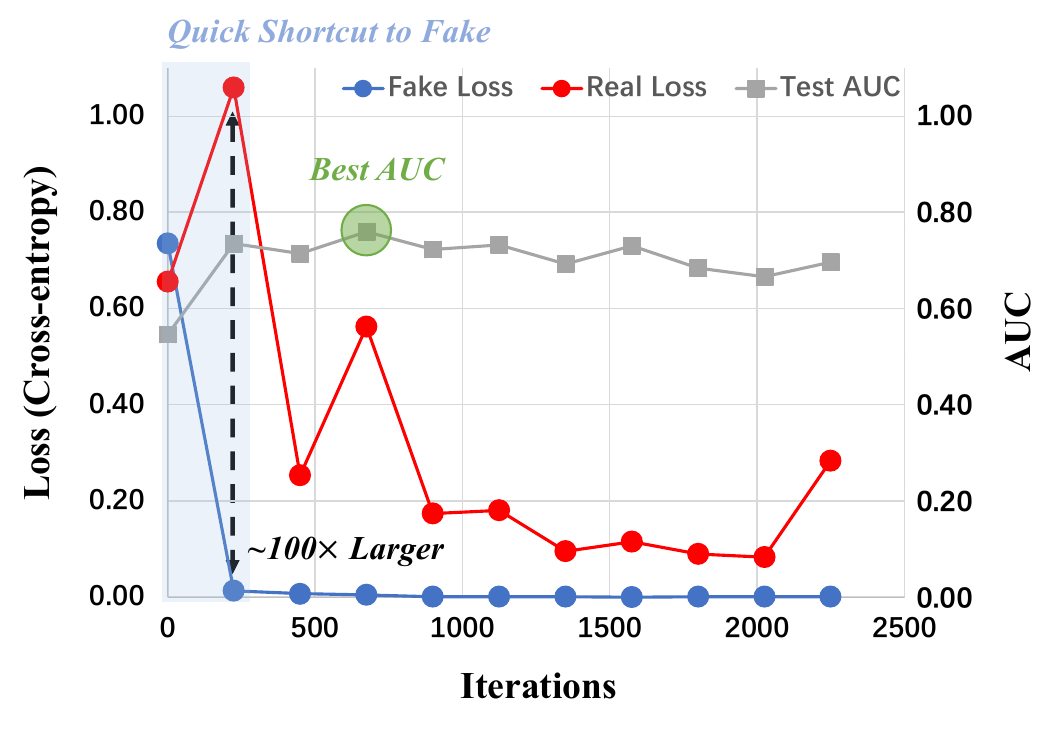}
    \vspace{-1.5em}
    \caption{\textbf{Illustration of the \textbf{asymmetry phenomenon} in AI-generated image detection}. 
    We show that the baseline detector (\textit{i.e.,} Xception) tends to \textbf{quickly overfit to the fake patterns} in the training set~\cite{rossler2019faceforensics++}, causing the limited generalization when facing previously unseen fakes~\cite{li2019celeb}.}
\label{fig:first_impression}
\vspace{-5mm}
\end{figure}

\vspace{-4mm}

\section{Introduction}
The rapid development of AI generative technologies has significantly lowered the barrier for creating highly realistic fake images. As deep generative models advance and mature~\cite{goodfellow2020generative,ddpm,rombach2022high,yan2025gpt}, the proliferation of AI-generated images (AIGIs\footnote{In the context of this research, AIGI primarily refers to deepfakes (face-swapping) and synthetic images (\textit{e.g.,} nature or arts).}) has drawn considerable attention, driven by their ability to produce high-quality content with relative ease. However, these advancements also introduce significant risks, if misused for malicious purposes such as deepfakes (mainly including face-swapping~\cite{korshunov2018deepfakes} and face-reenactment~\cite{thies2016face2face}), which may violate personal privacy, spread misinformation, and erode trust in digital media. Consequently, there is an urgent need to develop a reliable and robust framework for detecting AIGIs.

\begin{figure}
    \centering
    \includegraphics[width=1\linewidth]{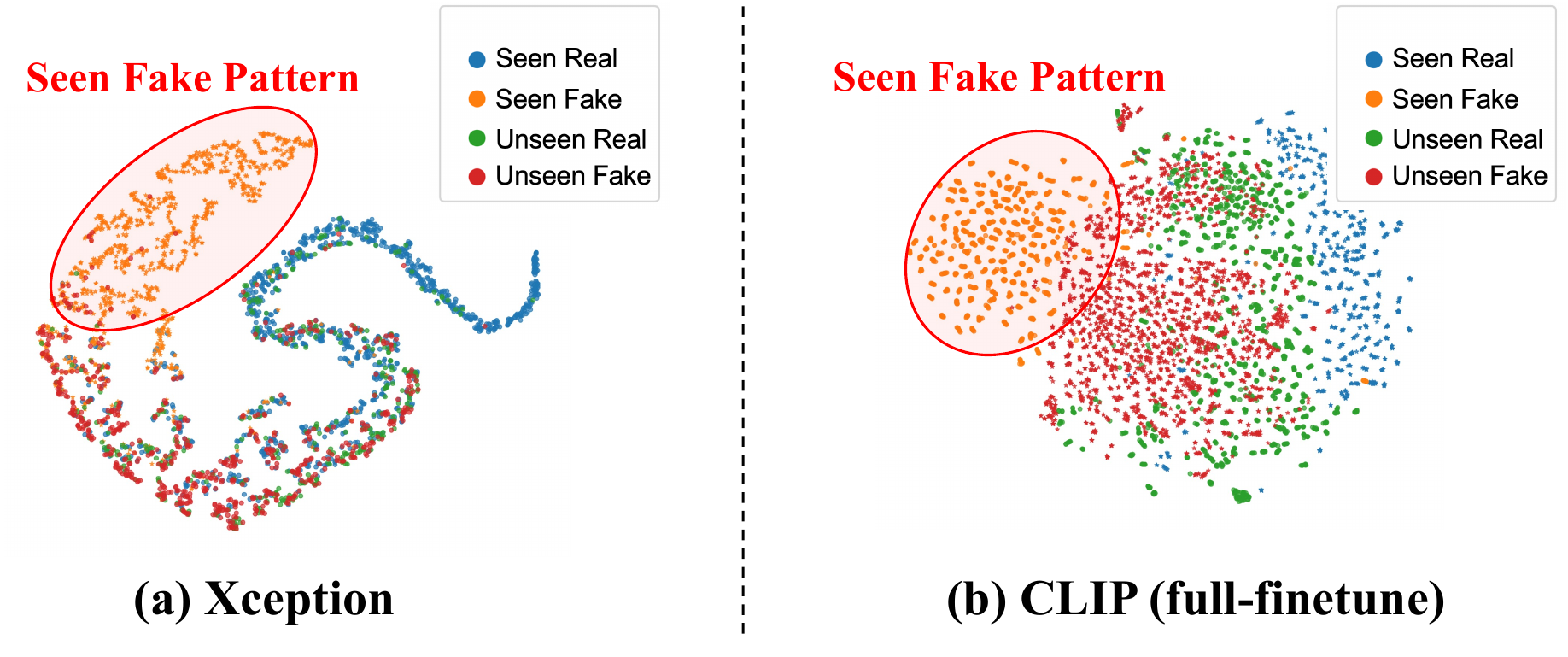}
    \vspace{-6mm}
    \caption{\textbf{t-SNE visualizations between Xception and CLIP (full-finetune)}.
    We show that both models only learn the specific fake patterns within the training set, treating samples with seen fake patterns as fake while other samples are all considered real, thereby limiting their generalization in detecting unseen fakes.
    }
\label{fig:comp_tsne}
\vspace{-5mm}
\end{figure}

Most existing studies in AIGI detection \cite{wang2020cnn, rossler2019faceforensics++} typically approach the real/fake classification problem as a symmetric binary classification task, akin to the ``cat versus dog" problem. A standard binary classifier, often based on deep neural networks, is trained to distinguish between real and fake images by predicting the likelihood of a given test image being fake during inference. Although this paradigm yields promising results when the training and testing distributions (in terms of fake generation methods) are similar, its performance tends to degrade significantly when applied to previously unseen fake methods, indicating the generalization issue~\cite{ojha2023towards}.

To understand the \textit{underlying reasons} for the failure in generalization, we have conducted extensive preliminary investigations and identified an \textit{asymmetry phenomenon} in AIGI detection: naively trained detectors tend to take the shortcut and very quickly overfit the limited fake patterns presented in the training set. Visualization in Fig.~\ref{fig:first_impression} corroborates this claim. Specifically, the vanilla detector (\textit{i.e.,} Xception~\cite{rossler2019faceforensics++}) quickly fits the fake patterns at the very early training stage (only a few iterations), resulting in a very low loss of fake, while the real loss is significantly higher than the fake loss ($\sim$ 100$\times$ larger).
This is likely because existing AIGI detection datasets~\cite{rossler2019faceforensics++,wang2020cnn} typically contain limited and homogeneous fake types, while real samples exhibit significantly greater diversity and variance between each other such as different categories and scenarios.

Consequently, the learned feature spaces become \textbf{fake-dominated and thus highly constrained.}
As evidenced by the t-SNE visualization in Fig.~\ref{fig:comp_tsne}, we see that the whole feature space is indeed dominated by the forgery patterns, where both the Xception (Vanilla CNN) and CLIP~\cite{clip_paper} detector group only the specific fake patterns within the training set into a single cluster, while all other data, including real samples and fake samples from unseen forgeries, are mapped into a separate cluster.

\begin{figure}[t]
    \centering
    \includegraphics[width=1\linewidth]{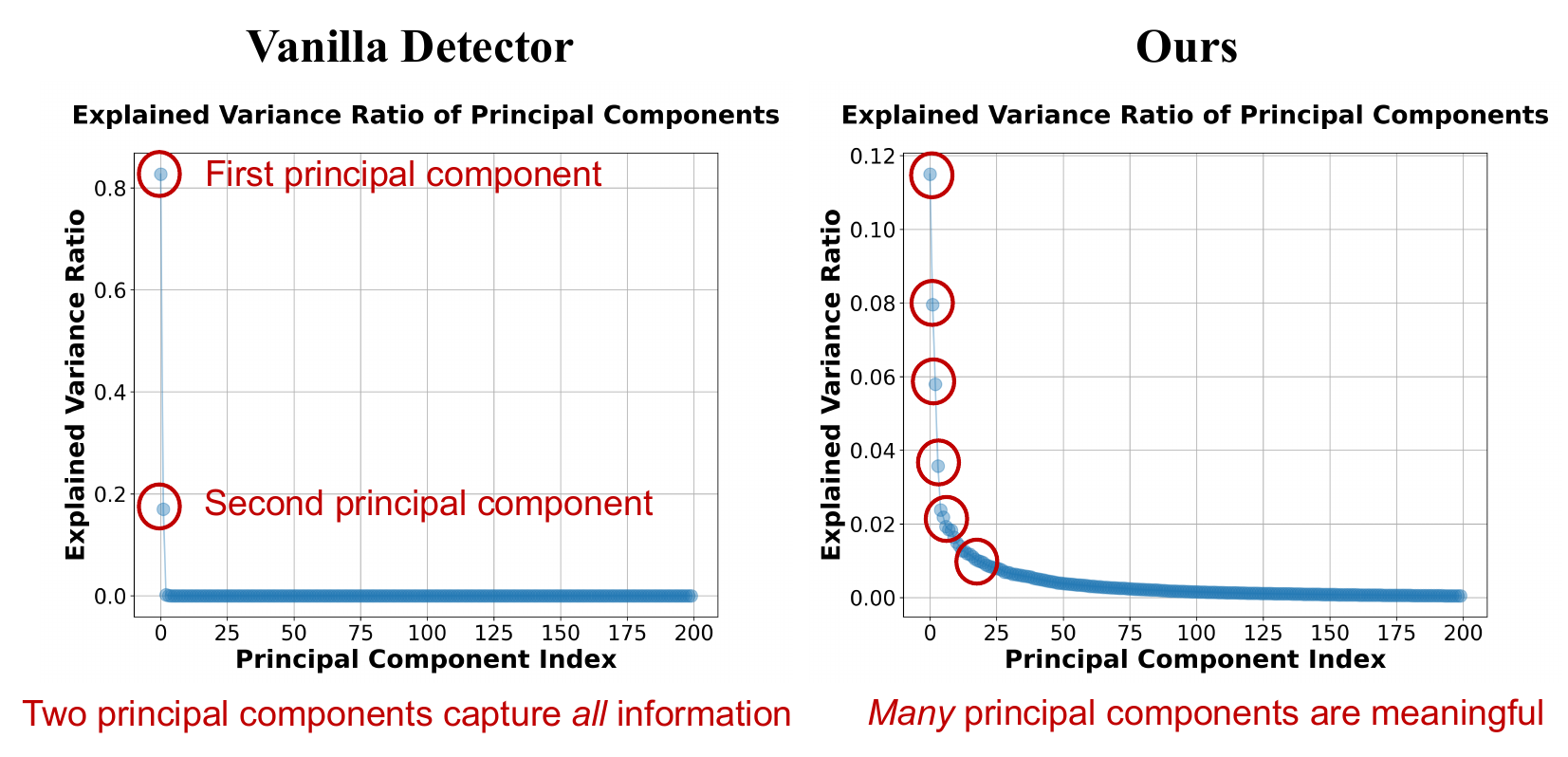}
    \vspace{-6mm}
    \caption{\textbf{Analysis for the \textit{effective information} contained in the model's feature space.} We apply PCA for dimension reduction and visualize the explained variance ratio of principal components with high contribution. We show that the baseline model trained on the AIGI dataset can be highly constrained and low-ranked. }
\label{fig:pca}
\vspace{-4mm}
\end{figure}

To quantify this, we analyze the \textit{effective information}\footnote{In PCA, it refers to variance captured by principal components. Larger eigenvalues indicate components explaining more variance and contributing more to data representation.} contained in the model's feature space via \textbf{Principal Component Analysis (PCA)}. Specifically, we visualize the \textit{explained variance ratio} of different principal components for the model's feature space in Fig.~\ref{fig:pca}. 
The results show that the feature space of the naive detector can be highly constrained and low-ranked, with \textbf{only two principal components to capture all information}, resulting in limited generalization.
This aligns with the previous theoretical analysis \cite{gunasekar2017implicit} that \textit{low-ranked\footnote{The ``rank" here means the number of significant principal components.} feature spaces hinder generalization by memorizing trivial patterns.}

One potential remedy to address the low-ranked problem is incorporating pre-trained knowledge within vision foundation models (VFMs), which provide higher-ranked feature representations, to \textbf{expand the low-ranked feature space}, thus alleviating the overfitting.
However, naively fine-tuning a VFM (even CLIP) risks distorting the original rich representation feature space (Fig.~\ref{fig:comp_tsne}), pushing the feature space to become low-ranked again (verified in Fig.~\ref{fig:pca_results}).

To address this, we design a novel approach called \textbf{Effort}: \textbf{Eff}icient \textbf{ort}hogonal modeling for generalizable AIGI detection. 
Specifically, we employ Singular Value Decomposition (SVD) to \textbf{construct two orthogonal subspaces}.
By freezing the principal components and adapting the remained components, we preserve the pre-trained knowledge while learning forgery-related patterns.

We have conducted extensive experiments on both deepfake detection and synthetic image detection benchmarks and find that our approach achieves significant superiority over other SOTAs with very little training cost.
Compared to existing full-parameters and LoRA-based tuning methods, we explicitly ensure orthogonality, enabling the higher rank of the whole feature space, effectively minimizing overfitting to fake and enhancing generalization.

\begin{figure}[!ht] 
\centering 
\includegraphics[width=0.5\textwidth]{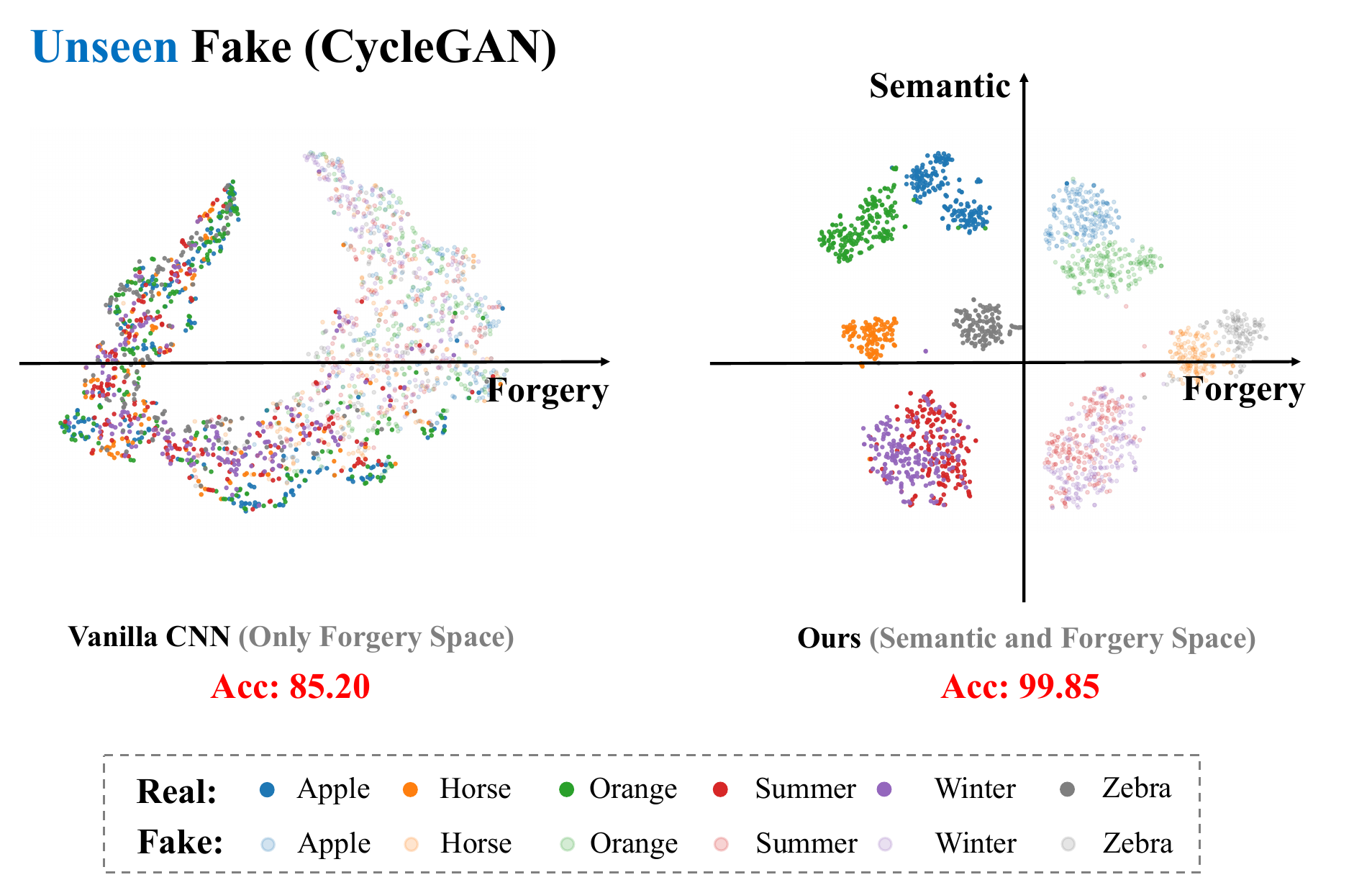} 
\vspace{-3mm}
\caption{
\textbf{t-SNE visualizations of the latent feature spaces between vanilla CNN~\cite{wang2020cnn} and ours.} Our method achieves orthogonal learning between the dimensions of semantic and forgery, allowing the model to capture fake patterns on the semantically-aligned subspace, simplifying the discrimination and thereby improving the generalization.
} 
\label{fig:semantic} 
\vspace{-4mm}
\end{figure}

Finally, we arrive at \textbf{a key insight for generalizable AIGI detection}: there exists an important prior of the detection task, where fake images are generated from real ones, establishing a \textbf{hierarchical relationship} rather than an independent or symmetric one. When aligning semantic information-such as distinguishing a fake dog from a real dog-this prior allows the model to focus discrimination within a smaller, semantically consistent subspace, e.g., only among dogs (see Fig.~\ref{fig:semantic} for illustration). This focused discrimination simplifies the task and aligns with theoretical results from Rademacher complexity~\cite{mohri2008rademacher}, which states that reducing model complexity leads to tighter generalization bounds. In contrast, naively trained detectors that treat real and fake data as independent fail to capture this structure, resulting in limited generalization performance. Therefore, modeling this hierarchical prior, we believe, is vitally crucial for AIGI detection.


Our work makes the following key contributions:
\vspace{-4mm}

\begin{itemize}
\item \textbf{Asymmetry phenomenon in AIGI detection}: We introduce the concept of \textit{asymmetry phenomenon}, where a naively trained detector tends to quickly fit the seen fake methods well but, in doing so, it often overfits to specific fake patterns in the training set, limiting its generalization ability to detect unseen fake methods.

\item \textbf{New perspective to explain the failure reasons behind generalization}: We use PCA to quantitatively assess the \textit{effective information} within the learned feature space, and we find that the overfitting to fake, results in a \textit{highly low-ranked and constrained feature space}, thus leading to the limited generalization capability.

\item \textbf{Novel method via orthogonal subspace decomposition}: To address the overfitting, we propose a novel approach, \textit{Effort}, with two careful designs: (1) incorporating the pre-trained knowledge (proving higher-ranked feature space) within the vision foundation models to expand the previous feature space, improving the model's expressivity and alleviating the overfitting; and (2) employing SVD to \textit{explicitly} construct two \textit{orthogonal subspaces}, where the principal one for preserving pre-trained knowledge and the remained one for learning new forgeries, avoiding the distortion of original rich feature space during learning fakes.

\item \textbf{Key insight toward generalizable detection}: We reveal that fake data is actually generated from real data, forming a hierarchical relationship rather than being independent. Our method effectively models this prior by maintaining the pre-trained semantic components while adapting to fake detection effectively, enabling the detector to make discrimination on the semantic-aligned subspaces, reducing model complexity and thus improving generalization.

\end{itemize}


\section{Related Work}
\label{sec:related}
Our work focuses on detecting AI-generated images (AIGIs), especially \textbf{deepfake images} (\textit{e.g.,} face-swapping) and \textbf{synthetic images} (\textit{e.g.,} nature or art), following \citet{yan2024transcending}.
As the majority of recent works specifically focus on dealing with the generalization issue, where the training and testing distribution differ (in terms of fake methods), we will briefly introduce the classical and recent detection methods toward generalization in deepfake and synthetic images, respectively.


\paragraph{Generalizable Deepfake Image Detection.} 
The task of deepfake detection grapples profoundly with the issue of generalization. 
To tackle the generalization issue, one mainstream approach is \textit{fake pattern learning}. Most existing works are within this line. These methods generally design a ``transformation function", \textit{e.g.,} frequency transformation~\cite{li2021frequency,luo2021generalizing,liu2021spatial}, blending operations~\cite{li2020face,zhao2021learning,shiohara2022detecting,chen2022self}, reconstruction~\cite{zhu2021face,cao2022end}, content/ID disentanglement~\cite{yan2023ucf,fu2025exploring,huang2023implicit,dong2023implicit}, to transform the original input $x$ into $x'$, where they believe that the more general fake patterns can be captured within the feature space of $x'$ compared to $x$. 
However, given the ever-increasing diversity of forgery methods in the real world, it is unrealistic to elaborate all possible fake patterns and ``expect" good generalization on unseen fake methods. 
Another notable direction is to \textit{real distribution learning}, with a specific methodology involved: one-class anomaly detection~\cite{khalid2020oc,larue2023seeable}.
Specifically, \citet{khalid2020oc} introduced a one-class-based anomaly detection, where ``abnormal" data is detected by the proposed reconstruction error as the anomaly score. \citet{larue2023seeable} proposed a similar approach to create pseudo-fake ``anomaly" samples by using image-level blending on different facial regions. 
However, it is challenging to ensure that the detector can learn a robust representation of real images by using the very limited real data in existing deepfake datasets (\textit{e.g.,} the FF++ dataset~\cite{rossler2019faceforensics++} contains only 1,000 real videos with imbalanced facial attribute distributions~\cite{trinh2021examination}).

\vspace{-4mm}

\paragraph{Generalizable Synthetic Image Detection.} 
With the rapid advancement of existing AI generative technologies, the scope of forged content has expanded beyond facial forgeries to encompass a wide range of scenes.
In this context, similar to the deepfake detection field, most existing works typically focus on \textit{fake pattern learning} that mines the low-level forgery clues from different aspects.
Specifically, several approaches have been proposed to capture low-level artifacts, including RGB data augmentations~\cite{wang2020cnn}, frequency-based features \cite{jeong2022bihpf}, gradients \cite{Tan_2023_CVPR}, reconstruction artifacts \cite{wang2023dire,chendrct,luo2024lare}, and neighboring pixel relationships \cite{tan2024rethinking}, random-mapping feature \cite{tan2024data}. 
To illustrate, BiHPF \cite{jeong2022bihpf} amplifies artifact magnitudes through the application of dual high-pass filters, while LGrad \cite{Tan_2023_CVPR} uses gradient information from pre-trained models as artifact representations. NPR \cite{tan2024rethinking} introduces a straightforward yet effective artifact representation by rethinking up-sampling operations. 
In addition to learning from scratch, there are also several research works~\cite{ojha2023towards,wu2023generalizable,liu2024forgery} that perform \textit{fake pattern learning} by leveraging the existing vision foundation models.
For instance, UniFD \cite{ojha2023towards} directly freezes the visual encoder of the pre-trained CLIP model and tunes only a linear layer for binary classification, demonstrating effective deepfake detection even with previously unseen sources. 
LASTED \cite{wu2023generalizable} proposes designing textual labels to supervise the CLIP vision model through image-text contrastive learning, advancing the field of synthetic image detection.
These arts have shown notable improvement in generalization performance when facing previously unseen fake methods.


\begin{figure*}[t]
    \centering
    \includegraphics[width=0.95\linewidth]{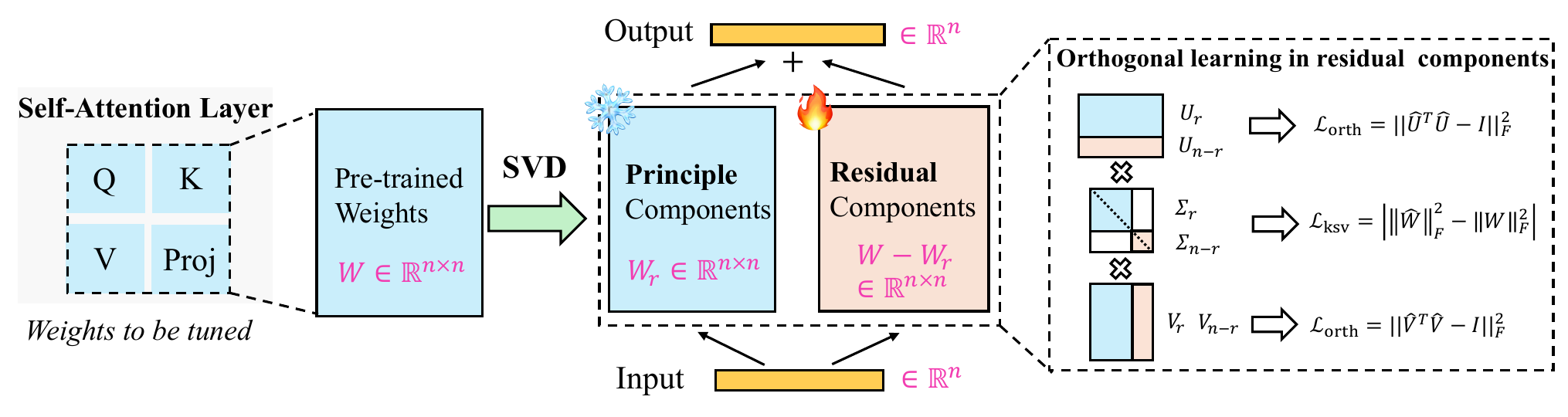}
    \vspace{-1em}
    \caption{\textbf{The proposed approach for AIGI detection}. The left branch is the decomposition matrix of the principle components approximation using SVD, while the right residual branch enables the orthogonal learning of real/fake discriminative features. 
    }
\label{fig:easy_framework}
\vspace{-3mm}
\end{figure*}

\section{Methodology} 
\label{framework}

The overall pipeline of the proposed \textit{Effort} approach is illustrated in Fig.~\ref{fig:easy_framework}, aiming to address the asymmetry phenomena in AIGI detection. 
Our approach involves the SVD to construct explicit orthogonality for preserving pre-trained knowledge and learning forgery-related patterns, avoiding the distortion of well-learned pre-trained knowledge during learning forgeries.

Formally, given a pre-trained weight matrix $W \in \mathbb{R}^{d_1 \times d_2}$ for a certain linear layer, we perform SVD to decompose $W$: \begin{equation} W = U \Sigma V^\top, \end{equation} where $U \in \mathbb{R}^{d_1 \times d_1}$ and $V \in \mathbb{R}^{d_2 \times d_2}$ are orthogonal matrices containing the left and right singular vectors, respectively, and $\Sigma \in \mathbb{R}^{d_1 \times d_2}$ is a diagonal matrix with singular values on the diagonal. Since the linear layer of VFM generally has the same input and output dimensions, we consider the case of SVD with $d_1 = d_2 = n$ in the following discussion.

To obtain a rank-$r$ approximation of the pre-trained weight matrix, we retain only the top $r$ singular values and corresponding singular vectors: \begin{equation} W \approx W_r = U_r \Sigma_r V_r^\top, \end{equation} where $U_r \in \mathbb{R}^{n \times r}$, $\Sigma_r \in \mathbb{R}^{r \times r}$, and $V_r \in \mathbb{R}^{n \times r}$. We keep $W_r$ frozen during training to preserve dominant pre-trained knowledge learned from large-scale data.

The residual component, defined as the difference between the pre-trained weights and the SVD approximation, is used to learn representations specific to fake images: \begin{equation} \Delta W = W - W_r = U_{n-r} \Sigma_{n-r} V_{n-r}^\top, \end{equation}
where $U_{n-r} \in \mathbb{R}^{n \times (n-r)}$, $\Sigma_{n-r} \in \mathbb{R}^{(n-r) \times (n-r)}$, $V_{n-r} \in \mathbb{R}^{n \times (n-r)}$. It is important to note that $\Delta W$ represents a learnable form associated with the remaining singular value decomposition, reflecting slight modifications or perturbations to the original weight matrix.

During training, we only optimize $\Delta W$ while keeping $U_r$, $\Sigma_r$, and $V_r$ fixed. This implementation ensures that the model retains its capability to process real images via the SVD approximation and adapts to detect deepfakes through the trivial residual components of the weight matrix.

To encourage the $\Delta W$ to capture both useful and meaningful discrepancy between the real and fake, it's significant to guarantee that optimizing $\Delta W$ does not change the properties of the overall weight matrix $W$ (\textit{i.e.}, Minimize the impact on the real information of the pre-trained weight as much as possible). Thus, we proposed two constraints to realize this goal, as follows.

\noindent\textbf{Orthogonal Constraint.} We maintain the orthogonality among each singular vector to keep orthogonal subspace for learning real/fake: \begin{equation} \mathcal{L}_{\rm orth} = \lVert \hat{U}^\top \hat{U} - I \rVert_F^2 + \lVert \hat{V}^\top \hat{V} - I \rVert_F^2, \end{equation} where $\hat{U} \in \mathbb{R}^{n \times n}$ denote the concatenation of $U_r$ and $U_{n-r}$ along the row dimension, $\hat{V} \in \mathbb{R}^{n \times n}$ denote the concatenation of $V_r$ and $V_{n-r}$ along the row dimension, $\lVert \cdot \rVert_F$ denotes the Frobenius norm, and $I$ is the identity matrix of appropriate dimensions.

\noindent\textbf{Singular Value Constraint.} The singular values can be interpreted as a type of scaling that affects the magnitude of the corresponding singular vectors. There is a relationship between singular values and the Frobenius norm of the weight matrix being decomposed: \begin{equation} \lVert W \rVert_F = \sqrt{\sum_i \sigma_i^2}, \end{equation} where $\sigma_i$ denotes the $i$-th singular value of the corresponding weight matrix.

To maximize the reduction of the impact of real knowledge, we constrain the singular values of the optimized weight matrix $\hat{W}$ to remain consistent with those of the original weight matrix $W$: 
\begin{equation}
\mathcal{L}_{\rm ksv} = \bigg| \sum_{i=r+1}^n \hat{\sigma}_i^2 - \sum_{i=r+1}^n \sigma_i^2 \bigg| = \bigg| \lVert \hat{W} \rVert_F^2 - \lVert W \rVert_F^2 \bigg|,
\end{equation}
where $\hat{W}$ represents the weights after the optimization of $W$, and $\left| \cdot \right|$ represents the absolute value. Note that this regularization will control the importance of the $\Delta W$ during optimization to prevent overfitting of learning real/fake.

\noindent\textbf{Loss Function.} The overall loss function for training the model combines the classification loss $\mathcal{L}_{\rm cls}$ (e.g., cross-entropy loss for binary classification) and the orthogonality regularization loss: \begin{equation} \mathcal{L} = \mathcal{L}_{\rm cls} + \lambda_1 \frac{1}{m}\sum_i^m \mathcal{L}_{\rm orth}^i + \lambda_2 \frac{1}{m}\sum_i^m \mathcal{L}_{\rm ksv}^i, \end{equation} where $\lambda_1, \lambda_2$ are hyperparameters that balance the importance of the corresponding regularization term, and $m$ represents the number of pre-trained weight matrices on which our approach is applied.
In practice, we adapt our approach to the \textit{linear layers} within the self-attention module across all transformer layers of the VFM to leverage their rich, well-learned real distributions.

Finally, we provide an algorithm illustration of the proposed approach in Alg.~\ref{alg:effort} for an overall understanding.

\vspace{-2mm}

\begin{algorithm}[!ht]
\small
\caption{\textit{Effort} Approach Algorithm}
\label{alg:effort}

\renewcommand{\algorithmicrequire}{\textbf{Input:}}
\renewcommand{\algorithmicensure}{\textbf{Output:}}
\begin{algorithmic}[1]
\REQUIRE Pre-trained weight matrix $W \in \mathbb{R}^{n \times n}$; Rank $r$; Training data $\mathcal{D} = \{(x_i, y_i)\}^{N}_{i=1}$; Hyperparameters $\lambda_1$, $\lambda_2$\\
\ENSURE Updated weight matrix $W$\\

\STATE \textit{$\triangleright$} \textbf{Step 1: Singular Value Decomposition}
\STATE Decompose $W$ via SVD: $W = U \Sigma V^\top$
\STATE Retain top $r$ singular values and vectors:
\STATE \quad $U_r \in \mathbb{R}^{n \times r}$, $\Sigma_r \in \mathbb{R}^{r \times r}$, $V_r \in \mathbb{R}^{n \times r}$
\STATE Compute $W_r = U_r \Sigma_r V_r^\top$
\STATE Keep $W_r$ fixed during training
\STATE Compute residual component: $\Delta W = W - W_r$
\STATE Decompose $\Delta W$ via SVD: $\Delta W = U_{n - r} \Sigma_{n - r} V_{n - r}^\top$
\STATE Initialize $\Delta W$
\STATE Define concatenated matrices:
\STATE \quad $\hat{U} = [U_r, U_{n - r}] \in \mathbb{R}^{n \times n}$
\STATE \quad $\hat{V} = [V_r, V_{n - r}] \in \mathbb{R}^{n \times n}$

\STATE \textit{$\triangleright$} \textbf{Step 2: Training Loop}
\FOR{each epoch}
    \FOR{each batch in $\mathcal{D}$}
        \STATE \textit{$\triangleright$} Forward Pass
        \STATE Compute model output using $W = W_r + \Delta W$
        \STATE Compute classification loss $\mathcal{L}_{\rm cls}$
        
        \STATE \textit{$\triangleright$} Compute Constraints
        \STATE Compute orthogonality loss:
        \STATE \quad $\mathcal{L}_{\rm orth} = \left\| \hat{U}^\top \hat{U} - I \right\|_F^2 + \left\| \hat{V}^\top \hat{V} - I \right\|_F^2$
        \STATE Compute singular value constraint loss:
        \STATE \quad $\mathcal{L}_{\rm ksv} = \left| \left\| \hat{W} \right\|_F^2 - \left\| W \right\|_F^2 \right|$
        
        \STATE \textit{$\triangleright$} Total Loss
        \STATE $\mathcal{L} = \mathcal{L}_{\rm cls} + \lambda_1 \mathcal{L}_{\rm orth} + \lambda_2 \mathcal{L}_{\rm ksv}$
        
        \STATE \textit{$\triangleright$} Backward Pass and Optimization
        \STATE Update $\Delta W$ using gradient descent to minimize $\mathcal{L}$
    \ENDFOR
\ENDFOR

\STATE \textbf{return} Updated weight matrix $W \gets W_r + \Delta W$
\end{algorithmic}
\end{algorithm}

\vspace{-3mm}

\section{Experiment}
\vspace{-1mm}

\begin{table*}
\centering
\caption{\textbf{Benchmarking Results of Cross-dataset Evaluations (Protocol-1) and Cross-method Evaluations (Protocol-2)}. 
All detectors are trained on FF++\_c23~\cite{rossler2019faceforensics++} and evaluated on other fake data. $\dagger$ indicates the results are obtained by using the model's checkpoint provided by the authors, otherwise, the results are cited from \cite{DeepfakeBench_YAN_NEURIPS2023,yan2024transcending,cheng2024can}.
}
\scalebox{0.53}{
\begin{tabular}{c|c|cccccccc|ccccccccc}
\toprule
\multirow{2}{*}{Methods}& Trainable & \multicolumn{8}{c|}{Cross-dataset Evaluation} & \multicolumn{9}{c}{Cross-method Evaluation} \\
\cmidrule(lr){3-10} \cmidrule(lr){11-19}
 & Param.& CDF-v2 & DFD & DFDC & DFDCP & DFo & WDF & FFIW & Avg. & UniFace & BleFace & MobSwap & e4s & FaceDan & FSGAN & InSwap & SimSwap & Avg. \\
\midrule
\midrule
F3Net~\cite{qian2020thinking} & 22M & 0.789 & 0.844 & 0.718 & 0.749 & 0.730 & 0.728 & 0.649 & 0.743 & 0.809 & 0.808 & 0.867 & 0.494 & 0.717 & 0.845 & 0.757 & 0.674 & 0.746 \\
SPSL~\cite{liu2021spatial} & 21M & 0.799 & 0.871 & 0.724 & 0.770 & 0.723 & 0.702 & 0.794 & 0.769 & 0.747 & 0.748 & 0.885 & 0.514 & 0.666 & 0.812 & 0.643 & 0.665 & 0.710 \\
SRM~\cite{luo2021generalizing} & 55M & 0.840 & 0.885 & 0.695 & 0.728 & 0.722 & 0.702 & 0.794 & 0.767 & 0.749 & 0.704 & 0.779 & 0.704 & 0.659 & 0.772 & 0.793 & 0.694 & 0.732 \\
CORE~\cite{ni2022core} & 22M & 0.809 & 0.882 & 0.721 & 0.720 & 0.765 & 0.724 & 0.710 & 0.762 & 0.871 & 0.843 & 0.959 & 0.679 & 0.774 & 0.958 & 0.855 & 0.724 & 0.833 \\
RECCE~\cite{cao2022end} & 48M & 0.823 & 0.891 & 0.696 & 0.734 & 0.784 & 0.756 & 0.711 & 0.779 & 0.898 & 0.832 & 0.925 & 0.683 & 0.848 & 0.949 & 0.848 & 0.768 & 0.844 \\
SLADD~\cite{chen2022self} & 21M & 0.837 & 0.904 & 0.772 & 0.756 & 0.800 & 0.690 & 0.683 & 0.777 & 0.878 & 0.882 & 0.954 & 0.765 & 0.825 & 0.943 & 0.879 & 0.794 & 0.865 \\
SBI~\cite{shiohara2022detecting} & 18M & 0.886 & 0.827 & 0.717 & 0.848 & 0.899 & 0.703 & 0.866 & 0.821 & 0.724 & 0.891 & 0.952 & 0.750 & 0.594 & 0.803 & 0.712 & 0.701 & 0.766 \\
UCF~\cite{yan2023ucf} & 47M & 0.837 & 0.867 & 0.742 & 0.770 & 0.808 & 0.774 & 0.697 & 0.785 & 0.831 & 0.827 & 0.950 & 0.731 & 0.862 & 0.937 & 0.809 & 0.647 & 0.824 \\
IID~\cite{huang2023implicit} & 66M & 0.838 & 0.939 & 0.700 & 0.689 & 0.808 & 0.666 & 0.762 & 0.789 & 0.839 & 0.789 & 0.888 & 0.766 & 0.844 & 0.927 & 0.789 & 0.644 & 0.811 \\
LSDA$\dagger$~\cite{yan2024transcending} & 133M & 0.875 & 0.881 & 0.701 & 0.812 & 0.768 & 0.797 & 0.724 & 0.794 & 0.872 & 0.875 & 0.930 & 0.694 & 0.721 & 0.939 & 0.855 & 0.793 & 0.835 \\
ProDet$\dagger$~\cite{cheng2024can} & 96M & 0.926 & 0.901 & 0.707 & 0.828 & 0.879 & 0.781 & 0.751 & 0.828 & 0.908 & 0.929 & 0.975 & 0.771 & 0.747 & 0.928 & 0.837 & 0.844 & 0.867 \\
CDFA$\dagger$~\cite{lin2024fake} & 87M & 0.938 & 0.954 & 0.830 & 0.881 & 0.973 & 0.796 & 0.777 & 0.878 & 0.762 & 0.756 & 0.823 & 0.631 & 0.803 & 0.942 & 0.772 & 0.757 & 0.781 \\
\midrule
\rowcolor{aliceblue} Effort (Ours) & 0.19M & 0.956 & 0.965 & 0.843 & 0.909 & 0.977 & 0.848 & 0.921 & 0.917 & 0.962 & 0.873 & 0.953 & 0.983 & 0.926 & 0.957 & 0.936 & 0.926 & 0.940 \\
\bottomrule
\end{tabular}
}
\vspace{-0.5em}
\label{tab:table1}
\end{table*}

\begin{table*}[!ht]
    \centering
    \caption{\textbf{ Benchmarking Results of Cross-method Evaluations in terms of AP Performance on the UniversalFakeDetect Dataset.}  $\dagger$ indicates that the results are obtained by using the official pre-trained model or reproduction. }
\resizebox{\textwidth}{20mm}{
    \begin{tabular}{c|c|c|c|c|c|c|c|c|c|c|c|c|c|c|c|c|c|c|c|c}
  \toprule
      \multirow{3}*{Methods}  & \multicolumn{6}{c|}{GAN} &  \multirow{3}*{\makecell[c]{Deep\\fakes}} & \multicolumn{2}{c|}{Low level} & \multicolumn{2}{c|}{Perceptual loss} &  \multirow{3}*{Guided} & \multicolumn{3}{c|}{LDM} & \multicolumn{3}{c|}{Glide} &  \multirow{3}*{Dalle}  &  \multirow{3}*{mAP}\\ 
      
       \cmidrule(lr){2-7} \cmidrule(lr){9-10}  \cmidrule(lr){11-12}  \cmidrule(lr){14-16}  \cmidrule(lr){17-19}  
       ~ & \makecell[c]{Pro-\\GAN} & \makecell[c]{Cycle-\\GAN} & \makecell[c]{Big-\\GAN} & \makecell[c]{Style-\\GAN}  & \makecell[c]{Gau-\\GAN}  & \makecell[c]{Star-\\GAN} & ~ & \makecell[c]{SITD}& \makecell[c]{SAN}& \makecell[c]{CRN}& \makecell[c]{IMLE}& ~ & {\makecell[c]{200\\steps}}& {\makecell[c]{200\\w/cfg}}& {\makecell[c]{100\\steps}}& {\makecell[c]{100\\27}} & {\makecell[c]{50\\27}} & \makecell[c]{100\\10} & ~ & ~\\
       \midrule
       \midrule
       CNN-Spot~\cite{wang2020cnn}  & 100.0 & 93.47 & 84.50 & 99.54 & 89.49 & 98.15 & 89.02 & 73.75 & 59.47 & 98.24 & 98.40 & 73.72 & 70.62 & 71.00 & 70.54 & 80.65 & 84.91 & 82.07 & 70.59 & 83.58 \\
       Patchfor~\cite{chai2020makes} & 80.88 & 72.84 & 71.66 & 85.75 & 65.99 & 69.25 & 76.55 & 76.19 & 76.34 & 74.52 & 68.52 & 75.03 & 87.10 & 86.72 & 86.40 & 85.37 & 83.73 & 78.38 & 75.67 & 77.73 \\
       Co-occurence~\cite{nataraj2019detecting} & 99.74 & 80.95 & 50.61 & 98.63 & 53.11 & 67.99 & 59.14 & 68.98 & 60.42 & 73.06 & 87.21 & 70.20 & 91.21 & 89.02 & 92.39 & 89.32 & 88.35 & 82.79 & 80.96 & 78.11 \\
       Freq-spec~\cite{zhang2019detecting} & 55.39 & 100.0 & 75.08 & 55.11 & 66.08 & 100.0 & 45.18 & 47.46 & 57.12 & 53.61 & 50.98 & 57.72 & 77.72 & 77.25 & 76.47 & 68.58 & 64.58 & 61.92 & 67.77 & 66.21 \\
       F3Net$\dagger$~\cite{qian2020thinking} & 99.96 & 84.32 & 69.90 & 99.72 & 56.71 & 100.0 & 78.82 & 52.89 & 46.70 & 63.39 & 64.37 & 70.53 & 73.76 & 81.66 & 74.62 & 89.81 & 91.04 & 90.86 & 71.84 & 76.89  \\
       UniFD~\cite{ojha2023towards} & 100.0 & 98.13 & 94.46 & 86.66 & 99.25 & 99.53 & 91.67 & 78.54 & 67.54 & 83.12 & 91.06 & 79.24 & 95.81 & 79.77 & 95.93 & 93.93 & 95.12 & 94.59 & 88.45 & 90.14 \\
       LGrad$\dagger$~\cite{Tan_2023_CVPR} & 100.0 & 93.98 & 90.69 & 99.86 & 79.36 & 99.98 & 67.91 & 59.42 & 51.42 & 63.52 & 69.61 & 87.06 & 99.03 & 99.16 & 99.18 & 93.23 & 95.10 & 94.93 & 97.23 & 86.35  \\
       FreqNet$\dagger$~\cite{tan2024frequency} & 99.92 & 99.63 & 96.05 & 99.89 & 99.71 & 98.63 & 99.92 & 94.42 & 74.59 & 80.10 & 75.70 & 96.27 & 96.06 & 100.0 & 62.34 & 99.80 & 99.78 & 96.39 & 77.78 & 91.95 \\ 
       NPR$\dagger$~\cite{tan2024rethinking} &  100.0 & 99.53 & 94.53 & 99.94 & 88.82 & 100.0 & 84.41 & 97.95 & 99.99 & 50.16 & 50.16 & 98.26 & 99.92 & 99.91 & 99.92 & 99.87 & 99.89 & 99.92 & 99.26 & 92.76  \\
       FatFormer$\dagger$~\cite{liu2024forgery} & 100.0 & 100.0 & 99.98 & 99.75 & 100.0 & 100.0 & 97.99 & 97.94 & 81.21 & 99.84 & 99.93 & 91.99 & 99.81 & 99.09 & 99.87 & 99.13 & 99.41 & 99.20 & 99.82 & 98.16 \\
      \midrule
       \rowcolor{aliceblue} Effort (Ours)  & 100.0 & 100.0 & 99.99 & 99.77 & 100.0 & 100.0 & 98.95 & 97.53 & 97.53 & 100.0 & 100.0 & 95.39 & 99.99 & 99.89 & 100.0 & 99.87 & 99.92 & 99.98 & 99.96 & 99.41 \\
\bottomrule
    \end{tabular}
}
\vspace{-0.5em}
  \label{tab:SOTA1}
\end{table*}

\begin{table*}[!ht]
    \centering
    \caption{\textbf{Benchmarking Results of Cross-method Evaluations in terms of Acc Performance on the UniversalFakeDetect Dataset.} $\dagger$ indicates that the results are obtained by using the official pre-trained model or reproduction. 
  }
\resizebox{\textwidth}{20mm}{
    \begin{tabular}{c|c|c|c|c|c|c|c|c|c|c|c|c|c|c|c|c|c|c|c|c}
    \toprule
      \multirow{3}*{Methods}  & \multicolumn{6}{c|}{GAN} &  \multirow{3}*{\makecell[c]{Deep\\fakes}} & \multicolumn{2}{c|}{Low level} & \multicolumn{2}{c|}{Perceptual loss} &  \multirow{3}*{Guided} & \multicolumn{3}{c|}{LDM} & \multicolumn{3}{c|}{Glide} &  \multirow{3}*{Dalle}  &  \multirow{3}*{mAcc}\\ 
      
       \cmidrule(lr){2-7} \cmidrule(lr){9-10}  \cmidrule(lr){11-12}  \cmidrule(lr){14-16}  \cmidrule(lr){17-19}  
       ~ & \makecell[c]{Pro-\\GAN} & \makecell[c]{Cycle-\\GAN} & \makecell[c]{Big-\\GAN} & \makecell[c]{Style-\\GAN}  & \makecell[c]{Gau-\\GAN}  & \makecell[c]{Star-\\GAN} & ~ & \makecell[c]{SITD}& \makecell[c]{SAN}& \makecell[c]{CRN}& \makecell[c]{IMLE}& ~ & {\makecell[c]{200\\steps}}& {\makecell[c]{200\\w/cfg}}& {\makecell[c]{100\\steps}}& {\makecell[c]{100\\27}} & {\makecell[c]{50\\27}} & \makecell[c]{100\\10} & ~ & ~\\
       \midrule
        \midrule
       CNN-Spot~\cite{wang2020cnn} & 99.99 & 85.20 & 70.20 & 85.70 & 78.95 & 91.70 & 53.47 & 66.67 & 48.69 & 86.31 & 86.26 & 60.07 & 54.03 & 54.96 & 54.14 & 60.78 & 63.80 & 65.66 & 55.58 & 69.58 \\
       Patchfor~\cite{chai2020makes} & 75.03 & 68.97 & 68.47 & 79.16 & 64.23 & 63.94 & 75.54 & 75.14 & 75.28 & 72.33 & 55.30 & 67.41 & 76.50 & 76.10 & 75.77 & 74.81 & 73.28 & 68.52 & 67.91 & 71.24 \\
       Co-occurence~\cite{nataraj2019detecting} &  97.70 & 63.15 & 53.75 & 92.50 & 51.10 & 54.70 & 57.10 & 63.06 & 55.85 & 65.65 & 65.80 & 60.50 & 70.70 & 70.55 & 71.00 & 70.25 & 69.60 & 69.90 & 67.55 & 66.86 \\
       Freq-spec~\cite{zhang2019detecting} & 49.90 & 99.90 & 50.50 & 49.90 & 50.30 & 99.70 & 50.10 & 50.00 & 48.00 & 50.60 & 50.10 & 50.90 & 50.40 & 50.40 & 50.30 & 51.70 & 51.40 & 50.40 & 50.00 & 55.45 \\
       F3Net$\dagger$~\cite{qian2020thinking} & 99.38 & 76.38 & 65.33 & 92.56 & 58.10 & 100.0 & 63.48 & 54.17 & 47.26 & 51.47 & 51.47 & 69.20 & 68.15 & 75.35 & 68.80 & 81.65 & 83.25 & 83.05 & 66.30 & 71.33  \\
       UniFD~\cite{ojha2023towards} & 100.0 & 98.50 & 94.50 & 82.00 & 99.50 & 97.00 & 66.60 & 63.00 & 57.50 & 59.50 & 72.00 & 70.03 & 94.19 & 73.76 & 94.36 & 79.07 & 79.85 & 78.14 & 86.78 & 81.38 \\
       LGrad$\dagger$~\cite{Tan_2023_CVPR} & 99.84 & 85.39 & 82.88 & 94.83 & 72.45 & 99.62 & 58.00 & 62.50 & 50.00 & 50.74 & 50.78 & 77.50 & 94.20 & 95.85 & 94.80 & 87.40 & 90.70 & 89.55 & 88.35 & 80.28  \\
       FreqNet$\dagger$~\cite{tan2024frequency} & 97.90 & 95.84 & 90.45 & 97.55 & 90.24 & 93.41 & 97.40 & 88.92 & 59.04 & 71.92 & 67.35 & 86.70 & 84.55 & 99.58 & 65.56 & 85.69 & 97.40 & 88.15 & 59.06 & 85.09 \\ 
       NPR$\dagger$~\cite{tan2024rethinking} & 99.84 & 95.00 & 87.55 & 96.23 & 86.57 & 99.75 & 76.89 & 66.94 & 98.63 & 50.00 & 50.00 & 84.55 & 97.65 & 98.00 & 98.20 & 96.25 & 97.15 & 97.35 & 87.15 & 87.56 \\
       FatFormer$\dagger$~\cite{liu2024forgery} & 99.89 & 99.32 & 99.50 & 97.15 & 99.41 & 99.75 & 93.23 & 81.11 & 68.04 & 69.45 & 69.45 & 76.00 & 98.60 & 94.90 & 98.65 & 94.35 & 94.65 & 94.20 & 98.75 & 90.86 \\
       \midrule
       \rowcolor{aliceblue} Effort (Ours) & 100.0 & 99.85 & 99.60 & 95.05 & 99.60 & 100.0 & 87.60 & 92.50 & 81.50 & 98.90 & 98.90 & 69.15 & 99.30 & 96.80 & 99.45 & 97.45 & 97.80 & 97.15 & 98.05 & 95.19 \\
\bottomrule
    \end{tabular}
}
\vspace{-0.5em}
  \label{tab:SOTA2}

\end{table*}

\subsection{Deepfake Image Detection}

\vspace{-2mm}

\noindent \textbf{Implementation Details.}
We utilize CLIP ViT-L/14~\cite{clip_paper} as the default vision foundation model (VFM). We also investigate other VFMs in Tab.~\ref{tab:backbone}.
We follow the pre-processing and training pipeline and use the codebases of DeepfakeBench~\cite{DeepfakeBench_YAN_NEURIPS2023}.
Additionally, we sample 8 frames from each video for training and 32 frames for inference, following \cite{shiohara2022detecting}. We use the fixed learning rate of 2e-4 for training our approach and employ the Adam~\cite{kingma2014adam} for optimization. We set the batch size to 32 for both training and testing.
We also employ several widely used data augmentations, such as Gaussian Blur and Image Compression, following other existing works~\cite{yan2024transcending,shiohara2022detecting,cheng2024can}.
For the evaluation metric, we report the widely-used video-level Area Under the Curve (AUC) to compare our approach with other works, following \cite{lin2024fake,shiohara2022detecting}. We compute the average model's output probabilities of each video to obtain the video-level AUC.

\vspace{-3mm}

\paragraph{Evaluation Protocols and Dataset.}
We adopt two widely used and standard protocols for evaluation: \textbf{Protocol-1:} cross-dataset evaluation and \textbf{Protocol-2:} cross-manipulation evaluation within the FF++ data domain.
For \textbf{Protocol-1}, we conduct evaluations by training the models on FaceForensics++ (FF++)~\cite{rossler2019faceforensics++} and testing them on other seven deepfake detection datasets: Celeb-DF-v2 (CDF-v2)~\cite{li2019celeb}, DeepfakeDetection (DFD)~\cite{dfd}, Deepfake Detection Challenge (DFDC)~\cite{dfdc}, the preview version of DFDC (DFDCP)~\cite{dfdcp}, DeeperForensics (DFo)~\cite{jiang2020deeperforensics}, WildDeepfake (WDF)~\cite{zi2020wilddeepfake}, and FFIW~\cite{zhou2021face}.
Note that FF++ has three different compression versions and we adopt the c23 version for training all methods in our experiments, following most existing works~\cite{yan2024transcending}.
For \textbf{Protocol-2}, we evaluate the models on the latest deepfake dataset DF40~\cite{yan2024df40}, which contains the forgery data generated within the FF++ domain, ensuring the fake methods different while the data domains remain unchanged.

\vspace{-3mm}

\paragraph{Evaluation Benchmarking.}
To provide a comprehensive benchmark for comparison, we introduce \textbf{13 competitive detectors}, including several classical detection methods such as F3Net~\cite{qian2020thinking} (ECCV'20), SPSL~\cite{liu2021spatial} (CVPR'20), SRM~\cite{liu2021spatial} (CVPR'21), CORE~\cite{ni2022core} (CVPRW'22), RECCE~\cite{cao2022end} (CVPR'22), and SBI~\cite{shiohara2022detecting} (CVPR'22), and also several latest SOTA methods (after 2023), such as UCF~\cite{yan2023ucf} (ICCV'23), IID~\cite{huang2023implicit} (CVPR'23), TALL~\cite{xu2023tall} (ICCV'23), LSDA~\cite{yan2024transcending} (CVPR'24), ProDet~\cite{cheng2024can} (NeurIPS'24), and CFDA~\cite{lin2024fake} (ECCV'24).
All detectors are trained on FF++ (c23) and tested on other fake data.
Results in Tab.~\ref{tab:table1} demonstrate two notable advantages of our approach. \textbf{(1) generalizability}: we see that our approach consistently and largely outperforms other models across basically all tested scenarios, validating that our method is generalizable for detecting unseen fake data, even for the latest face-swapping techniques such as BleFace~\cite{blendface}.
\textbf{(2) efficiency}: it is worth noting that our method only needs 0.19M parameters for training to achieve superior generalization. As we can see most latest SOTA detectors such as LSDA and ProDet all use about 100M parameters for training, while we are about \textbf{1,000$\times$ smaller}.

\vspace{-1mm}

\subsection{Synthetic Image Detection}
\vspace{-1pt}

\paragraph{Evaluation Metrics.}

We follow existing works \cite{cnn-detect, ojha2023towards, liu2024forgery} for benchmarking and report both average precision (AP) and classification accuracy (Acc). For Acc, we set the classification threshold for each dataset to $0.5$ to ensure a fair comparison.

\vspace{-5mm}
\paragraph{UniversalFakeDetect Dateset.}
We adhere to the protocol outlined in \cite{cnn-detect,ojha2023towards} and utilize ProGAN's real and fake images as our training dataset, which includes 20 subsets of generated images. The evaluation set contains 19 subsets derived from different kinds of generative models, including ProGAN~\cite{progan}, CycleGAN~\cite{Zhu-2017-cycleGAN}, BigGAN~\cite{brock2018biggan}, StyleGAN~\cite{stylegan}, GauGAN~\cite{park2019SPADE}, StarGAN~\cite{choi2018stargan}, DeepFakes~\cite{roessler2019faceforensicspp}, SITD~\cite{sitd}, SAN~\cite{san}, CRN~\cite{crn}, IMLE~\cite{imle}, Guided (guided diffusion model)\cite{guided-diffusion}, LDM (latent diffusion model) \cite{ldm}, Glide \cite{glide}, and DALLE \cite{dalle-orig}.
\vspace{-5mm}
\paragraph{Implementation Details.} 
Similar to the setting of deepfake image detection, we adopt pre-trained CLIP ViT-L/14 as the backbone and use the Adam optimizer \cite{kingma2014adam} with a fixed learning rate of 2e-4. 
The batch size is set to 48.
Other settings and details are the same with \cite{ojha2023towards}.

\begin{table}
  \centering
  \setlength{\tabcolsep}{12pt}
  \caption{\textbf{Ablation studies regarding the proposed SVD, singular value constraint ($\mathcal{L}_{\rm ksv}$), and orthogonal constraint ($\mathcal{L}_{\rm orth}$)}. 
  All models are trained on FF++ (c23) and tested on other datasets.
  }
  \label{tab:ortho_loss}
  \scalebox{0.71}{
  \begin{tabular}{c|c|c|c|c|c} 
  \toprule
    \multicolumn{3}{c|}{Ours} & \multirow{2}{*}{CDF-v2} & \multirow{2}{*}{SimSwap} & \multirow{2}{*}{Avg.} \\
    \cmidrule(r){1-3}
    \textit{SVD} & $\mathcal{L}_{\rm ksv}$ & $\mathcal{L}_{\rm orth}$ & & &  \\ 
    \midrule
     $\times$ & $\times$ & $\times$ &
     0.857 & 0.860 & 0.859 \\

     \checkmark & $\times$& $\times$ &
     0.940 & 0.910 & 0.925 \\

    \checkmark & \checkmark & $\times$ &
     0.944 & 0.927 & 0.936 \\

     \checkmark & $\times$ & \checkmark &
     0.945 & 0.914 & 0.930 \\

     \checkmark & \checkmark & \checkmark &
     0.956 & 0.926 & 0.941 \\
    
    \bottomrule
  \end{tabular}
  }
  \vspace{-3mm}
  
\end{table}

\begin{table}

  \centering
  \caption{\textbf{Ablation studies regarding different vision foundation models (VFMs) were used}. All models are trained on FF++ (c23) and tested on CDF-v2 and SimSwap.
  }
  \label{tab:backbone}
  \scalebox{0.65}{
  \begin{tabular}{c|c|c|c|c|c} 
  \toprule
    VFMs & \#Params & \#ImgSize & CDF-v2 & SimSwap & Avg. \\ 
    \midrule
    BEIT-v2~\cite{peng2022beit} & 303M & 224 & 0.855 & 0.821 & 0.838 \\
    + Ours & 0.14M & 224 & 0.894 & 0.850 & 0.872 \\ 
    \midrule
    SigLIP~\cite{zhai2023sigmoid} & 316M & 256 & 0.877 & 0.713 & 0.795 \\
    + Ours & 0.19M & 256 & 0.895 & 0.778 & 0.867 \\ 
    \midrule
    CLIP~\cite{clip_paper} & 307M & 224 & 0.857 & 0.860 & 0.859 \\
    + Ours & 0.19M & 224 & 0.956 & 0.926 & 0.941 \\ 
    \bottomrule
  \end{tabular}
  }
  \vspace{-5mm}
  
\end{table}

\paragraph{Evaluation Analysis.}
\label{sec:baselines}
\vspace{-1pt}
The AP and Acc results are presented in Tab.~\ref{tab:SOTA1} and Tab.~\ref{tab:SOTA2}, respectively. Our method attains impressive detection results, achieving 95.19\% mAcc and 99.41\% mAP across the 19 test subsets. 
One similar approach to ours is UniFD, which also preserves the original pre-trained knowledge of CLIP and fine-tunes only the FC layer for discrimination.
In contrast to UniFD which directly performs discrimination in the pre-trained knowledge space, our approach utilizes SVD to create an orthogonal low-ranked subspace for learning forgeries while preserving the essential high-ranked representational space, achieving the discrimination by leveraging both representational and forgery subspaces, achieving a significant improvement of 9.27\% in mAcc and 13.81\% in mAP over UniFD.
Besides, when compared to the SOTA method, FatFormer, we achieve 4.33\% mAcc improvement without relying on the extra text encoder of CLIP. This further demonstrates the superiority of our approach.

\begin{table}
  \centering
  \setlength{\tabcolsep}{10pt}
  \caption{\textbf{Ablation studies on synthetic image detection regarding the tunable $n-r$ values in SVD (Ours) and $r$ values in LoRA}. All models are trained on ProGAN's images and tested on 19 different generative models' images. ``FFT" indicates the full fine-tuning. ``Linear-Prob" indicates fine-tuning FC layer only, where we reproduce the results from UniFD~\cite{ojha2023towards}.
  }
  \resizebox{0.71\columnwidth}{!}{
  \begin{tabular}{c|c|c|c|c} 
    \toprule
    \multicolumn{2}{c|}{Archs.} & $n-r$ & $r$ & mAcc \\ 
    \midrule
    \multicolumn{2}{c|}{UniFD (Linear-Prob)} & -- & -- & 81.02 \\
    \multicolumn{2}{c|}{Baseline (FFT)} & -- & -- & 86.22 \\
    \midrule
    \multirow{5}{*}{LoRA} & Variant-1 & -- & 256 & 91.42 \\
    & Variant-2 & -- & 64 & 91.06 \\
    & Variant-3 & -- & 16 & 91.89 \\
    & Variant-4 & -- & 4 & 93.53 \\
    & Variant-5 & -- & 1 & 93.03 \\
    \midrule
    \multirow{5}{*}{Ours} & Variant-1 & 256 & -- & 92.13 \\
    & Variant-2 & 64 & -- & 93.68 \\
    & Variant-3 & 16 & -- & 94.45 \\
    & Variant-4 & 4 & -- & 94.37 \\
    & Variant-5 & 1 & -- & 95.19 \\
    \bottomrule
  \end{tabular}
  }
  \label{tab:lora_aigc}
  \vspace{-2mm}
\end{table}

\begin{table}
    \centering
    \caption{
        \textbf{Cross-dataset generalization evaluations with existing \textit{adapter-based} deepfake detectors}. The results are cited from their original papers. The metric is \textbf{frame-level AUC.}
    }
    \scalebox{0.74}{
        \begin{tabular}{c|c|c|c|c} \toprule
        Methods & CDF-v2 & DFD & DFDC & Avg.\\ 
        \midrule
        LoRA~\cite{kong2023enhancing} & 0.838 & 0.834 & 0.717 & 0.796 \\
        MoE-LoRA~\cite{kong2024moe} & 0.867 & 0.904 & -- & -- \\
        Dual-Adapter~\cite{shao2023deepfake} & 0.717 & -- & 0.727 & -- \\
        Ours & 0.901 & 0.923 & 0.798 & 0.874 \\
        \bottomrule
      \end{tabular}
    }
    \label{tab:cmp_with_lora}
    \vspace{-2mm}
\end{table}

\subsection{Ablation Study and Analysis}
\label{ablation_analysis}

\paragraph{Incremental improvement of the proposed designs.}
Ablation studies in Tab.~\ref{tab:ortho_loss} demonstrate incremental gains from the proposed SVD method and loss constraints: the baseline (no modules) achieves an average AUC of 0.859, while adding SVD alone boosts performance to 0.925 (+6.6\%), underscoring its efficacy in isolating forgery artifacts via orthogonal feature decomposition. Further integrating the singular value constraint ($\mathcal{L}_{\rm ksv}$) and orthogonal constraint ($\mathcal{L}_{\rm orth}$) refines performance, yielding 0.936 and 0.930 AUC, respectively, with their combined synergy (SVD + $\mathcal{L}_{\rm ksv}$ + $\mathcal{L}_{\rm orth}$) achieving peak performance (0.941 AUC). 
These results highlight the significant contribution of SVD as the primary driver of performance gains while illustrating the complementary benefits of $\mathcal{L}_{\rm ksv}$ and $\mathcal{L}_{\rm orth}$ in further refining the generalization performance.

\vspace{-2mm}

\paragraph{Compatibility with other vision foundation models.}
By default, we choose CLIP as the vision foundation model (VFM) in our experiments. To validate the generality and versatility of our approach, we conduct an ablation study to apply our method to other VFMs, including BEIT-v2~\cite{peng2022beit} and SigLIP~\cite{zhai2023sigmoid}.
Results in Tab.~\ref{tab:backbone} show that our approach can be seamlessly applied to other VFMs to improve the model's generalization performance.

\vspace{-2mm}

\begin{figure}[t]
    \centering
    \includegraphics[width=0.85\linewidth]{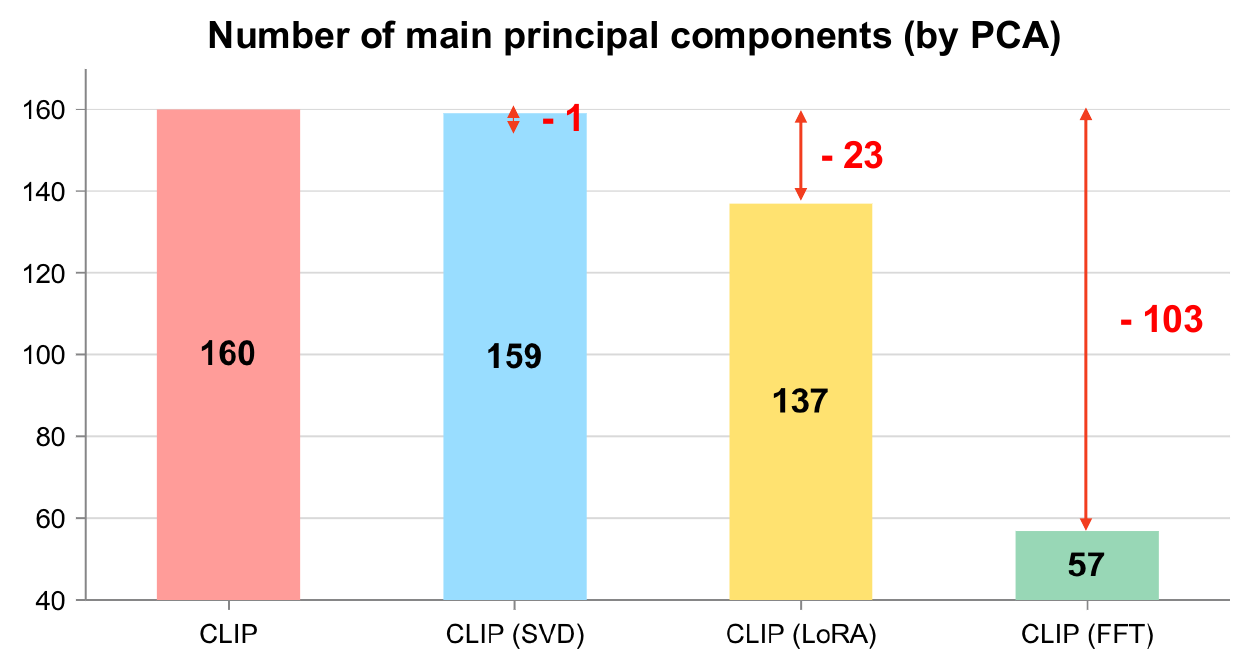}
    \vspace{-0.5em}
    \caption{
    \textbf{PCA quantifying feature space information retention, measured by the minimum number of components required to explain $\ge 90\%$ variance.}
    Our SVD-based fine-tuning preserves 159/160 principal components (retaining 99.3\% original variance), while LoRA and FFT exhibit notable degradation, with 14.4\% and 64.4\% variance loss, respectively. See Tab.~\ref{tab:rank_acc_comparison} for more results.
    }
\label{fig:pca_results}
\vspace{-4mm}
\end{figure}

\paragraph{Comparison with Existing Adapter-Based Detectors.}

Parameter-efficient fine-tuning (PEFT) has become a popular technique for adapting pre-trained large models to downstream tasks~\cite{ding2023parameter}.
Low-ranked adaptation (LoRA)~\cite{hu2021lora} is a widely used approach for PEFT. 
Previous works~\cite{kong2023enhancing,liu2024forgery} employing LoRA in the VFMs have achieved good empirical generalization results for detection. 
Additionally, \cite{kong2024moe} and \cite{shao2023deepfake} introduce MoE-LoRA techniques and dual adapters into the deepfake detection fields.
However, the existing adapter-based methods do \textbf{not explicitly ensure this orthogonality}, still having the potential to distort the pre-existing pre-trained knowledge and result in unexpected generalization results.
In contrast, our method \textbf{explicitly} constructs two \textbf{orthognoal subspaces} for pre-trained knowledge and forgery using SVD, ensuring the pre-existing pre-trained knowledge will not be distorted, thereby achieving better generalization performance. 
To verify this, we provide several empirical results in Tab.~\ref{tab:cmp_with_lora} and Tab.~\ref{tab:lora_aigc}.
From these results, we can see that our proposed SVD-based method achieves clearly higher generalization results than adapter-based methods in both deepfake detection and synthetic image detection fields, as our approach \textbf{explicitly preserves the pre-trained knowledge while learning the forgery patterns.} 
We also use PCA to compute the rank of the feature space, similar to Fig.~\ref{fig:pca}. Results in Fig.~\ref{fig:pca_results} highlight the superiority of our method, where full-parameters fine-tuning and LoRA can lead to a notable reduction (-103 and -23, respectively) while our method best retains the pre-trained knowledge, maintaining a higher-ranked feature space for better generalization.

\begin{table}
\centering
\caption{
\textbf{Comparison of different fine-tuning methods.} We compute both the effective rank and mean accuracy for evaluation. Our SVD retains most principal components (notably higher effective rank), thus achieving the best detection results.
}
\scalebox{0.65}{
\begin{tabular}{l|c|c|c|c} 
\toprule
\multirow{2}{*}{Fine-tuning Methods} & \multicolumn{2}{c|}{Face Domain} & \multicolumn{2}{c}{General Domain} \\
\cline{2-5}
& Effective Rank & Accuracy & Effective Rank & Accuracy \\
\midrule
Baseline (No FT) & 160 & - & 479 & - \\
SVD (ours) & 159 & 95.60 & 316 & 95.19 \\
LoRA & 137 & 89.40 & 304 & 93.03 \\
FFT & 57 & 85.70 & 238 & 86.22 \\
\bottomrule
\end{tabular}
}
\label{tab:rank_acc_comparison}
\vspace{-3mm}
\end{table}

\begin{figure}[t]
    \centering
    \includegraphics[width=1\linewidth]{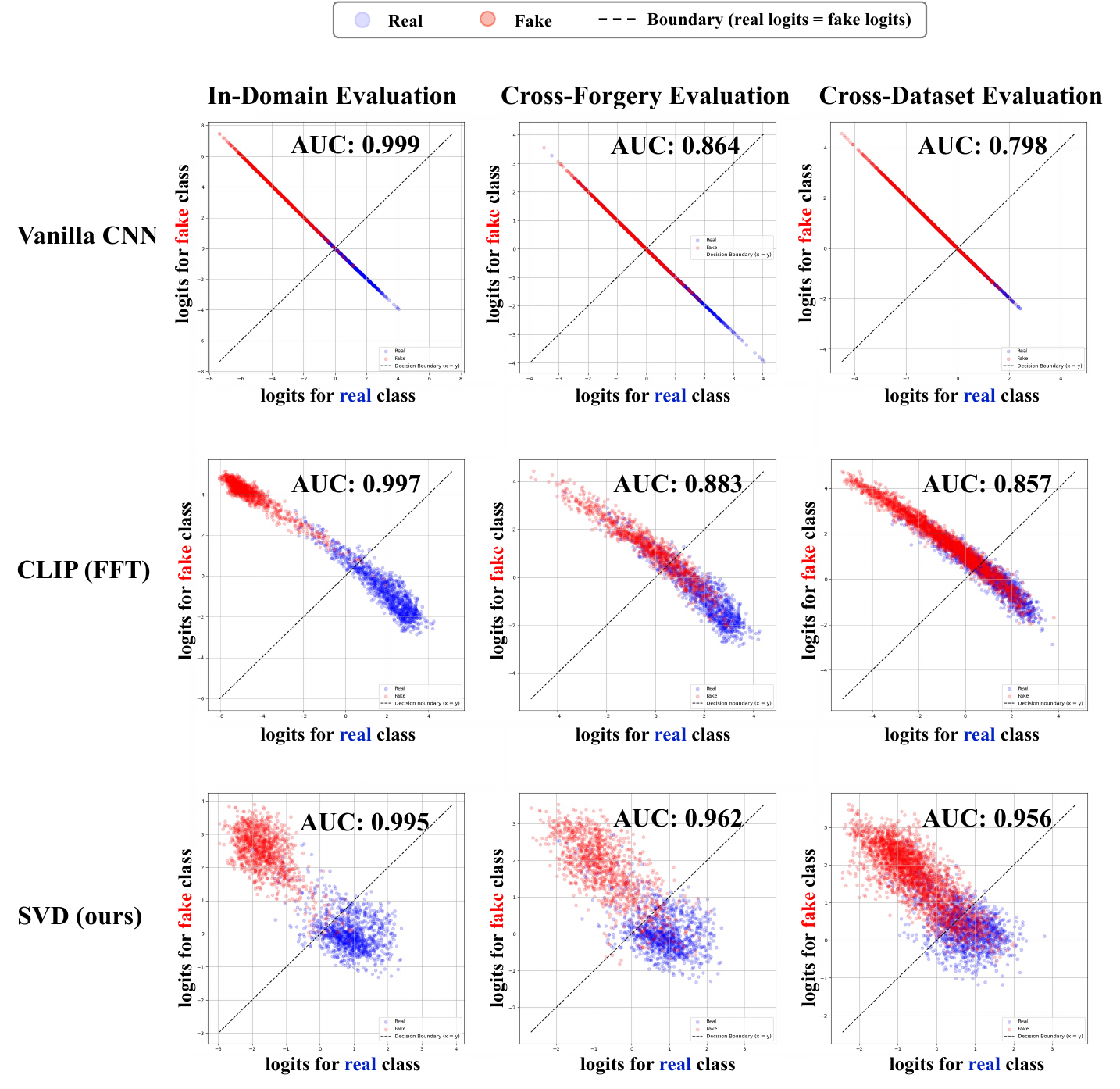}
    \vspace{-4mm}
    \caption{\textbf{Evidence for validating the discrimination dimension from the \textit{logits space}}.
    \textbf{The first row} shows that the vanilla CNN (\textit{i.e.,} Xception) overfits to the seen fakes and relies only on the forgery pattern for discrimination.
    \textbf{The second row} shows that fully fine-tuning the VFM (\textit{i.e.,} CLIP) involves pre-trained knowledge but also distorting part of it during learning forgery patterns. 
    \textbf{The third row} shows that our method uses SVD to achieve the orthogonality, thereby best retaining the pre-trained knowledge.
    }
\label{fig:real_fake_logits}
\vspace{-4mm}
\end{figure}

\vspace{-2mm}

\paragraph{The naively trained models leverage very limited forgery patterns for discrimination.}
In our motivation, we argue that conventional training paradigms cause models to over-rely on very limited forgery patterns for discrimination, thereby causing the highly low-ranked and constrained feature space, limiting their expressivity and generalization (Fig. \ref{fig:comp_tsne} and Fig. \ref{fig:pca}). 
To further validate this argument, we analyze the \textbf{discrimination behavior} of baseline models (\textit{e.g.,} Xception and CLIP) through decision boundary visualization using the model's output logits of real and fake classes (Fig. \ref{fig:real_fake_logits}). 
The linear alignment of predictions along $y= - x+b$ reveals that \textbf{Xception collapses real/fake discrimination into a single discriminative dimension}, confirming its exclusive dependence on forgery cues.
Notably, while CLIP's original higher-ranked representations initially preserve informative structures, full fine-tuning catastrophically degrades this structure, forcing decisions into a similarly collapsed subspace ($y= - x+b$). Our method addresses this concern by learning forgery-related features in a novel orthogonal complement subspace relative to CLIP's original representational embedding space, thereby achieving optimal generalization performance.

\paragraph{How to determine the value of rank in tuning SVD?}
Our choice of using a lower rank (specifically, ``n-r"=1) for fine-tuning DFD is primarily motivated by two critical factors:
\textbf{First, the nature of the real-fake classification task itself makes it relatively straightforward}. Specifically, fake samples in existing training sets tend to exhibit a limited number of distinctive forgery patterns (FF++ contains only four forgery types), each with relatively simple and consistent characteristics.
Due to this simplicity and limited diversity, a low-rank adaptation with a small rank (e.g., ``n-r"=1, 4, or 16) is sufficient for the model to effectively learn these forgery patterns. As demonstrated by Table 5 in our paper, choosing ranks of 1, 4, or 16 yields very similar performance results. Given this observation, we prioritize efficiency and parameter economy, making rank 1 the optimal choice.
\textbf{Second, the inherent characteristics of binary classification further justify selecting a smaller rank.} Binary classification tasks typically do not require the model to learn extensive and nuanced patterns, but rather to identify just enough distinctive features to separate the two classes, making the learned feature space inherently constrained. Thus, binary classification inherently simplifies the complexity of the learning problem, meaning that employing a higher rank would not provide significant additional benefit.

\vspace{-3mm}

\section{Conclusion}
\vspace{-1mm}

In this paper, we start our research from a new perspective to excavate the failure reason of the generalization in AIGI detection, namely the asymmetry phenomena, where a naively trained detector very quickly shortcuts to the seen fake patterns, collapsing the feature space into a low-ranked structure that limits expressivity and generalization. 
To address this, we propose integrating higher-ranked pre-trained knowledge from vision foundation models to expand the feature space. Simultaneously, we decompose the feature space into two orthogonal subspaces, for preserving pre-trained knowledge while learning forgery.
Beyond LoRA and full-parameters tuning, we explicitly ensure the orthogonality, maintaining the higher rank of the whole feature space for better generalization.
Furthermore, we reveal a very important prior for generalizable AIGI detection that fake data actually originates from real data in a hierarchical structure, not independently. Our method leverages this prior by preserving pre-trained semantic components while adapting to fake detection, enabling discrimination in semantic-aligned subspaces with reduced model complexity and improved generalization.
Extensive experiments with deep analysis on both deepfake and synthetic image detection benchmarks have demonstrated the superior advantages of the proposed method in AIGI detection.

\section{Impact Statements}
This research advances application-driven machine learning by proposing a novel method for detecting AI-generated images. By effectively identifying deepfakes and curbing malicious uses of generative models, it holds significant potential for positive social impact. However, a possible negative outcome is the misuse of our method to enhance the realism of deepfake generators. To address this concern, we plan to implement measures like access control. Overall, we urge the research community to minimize negative impacts while leveraging the positive contributions of this work.

\section{Acknowledgment}
This work was supported in part by the Natural Science Foundation of China (No. 62202014, 62332002, 62425101).

\clearpage
\bibliography{effort/main}

\clearpage
\appendix





\section{Additional Results and Ablations} \label{section2}

In this section, we provide additional experimental results and ablation studies of our proposed approach.

\begin{table*}[!ht]
    \centering
    \caption{\textbf{Benchmarking results of cross-method evaluations in terms of Acc performance on the Genimage dataset.} 
  We follow \cite{zhu2024genimage} and use the SDv1.4 as the training set while others as the testing sets. 
  We directly cite the results of ResNet-50, DeiT-S, Swin-T, CNNSpot, Spec, F3Net, and GramNet from \cite{zhu2024genimage}.
  We obtain the results of UnivFD and DRCT from \cite{chendrct}, and FreqNet, NPR, and FatFormer by using the official checkpoints for reproduction.
  We report the Accuracy metric for comparison following \cite{chendrct}.
  }
\resizebox{1.0\textwidth}{!}{
\setlength\tabcolsep{8pt}
    \begin{tabular}{c c c c c c c c c c | c}
    \bottomrule \hline
        Methods & Venues & Midjourney & SDv1.4 & SDv1.5 & ADM & GLIDE & Wukong & VQDM & BigGAN & mAcc\\
          \bottomrule \hline
ResNet-50~\cite{he2016deep}  &  CVPR 2016   & 54.9 & 99.9 & 99.7 & 53.5 & 61.9 & 98.2 & 56.6 & 52.0 & 72.1 \\
DeiT-S~\cite{touvron2021training}& ICML 2021 & 55.6 & 99.9 & 99.8 & 49.8 & 58.1 & 98.9 & 56.9 & 53.5 & 71.6 \\
Swin-T~\cite{liu2021swin}    &  ICCV 2021   & 62.1 & 99.9 & 99.8 & 49.8 & 67.6 & 99.1 & 62.3 & 57.6 & 74.8 \\
CNNSpot~\cite{wang2020cnn}    &  CVPR 2020  & 52.8 & 96.3 & 95.9 & 50.1 & 39.8 & 78.6 & 53.4 & 46.8 & 64.2 \\
Spec~\cite{zhang2019detecting} & WIFS 2019  & 52.0 & 99.4 & 99.2 & 49.7 & 49.8 & 94.8 & 55.6 & 49.8 & 68.8 \\
F3Net~\cite{qian2020thinking}  & ECCV 2020  & 50.1 & 99.9 & 99.9 & 49.9 & 50.0 & 99.9 & 49.9 & 49.9 & 68.7 \\
GramNet~\cite{liu2020global}   & CVPR 2020  & 54.2 & 99.2 & 99.1 & 50.3 & 54.6 & 98.9 & 50.8 & 51.7 & 69.9 \\
UnivFD~\cite{ojha2023towards}  & CVPR 2023  & 91.5&96.4&96.1&58.1&73.4&94.5&67.8&57.7 &79.5\\
NPR ~\cite{tan2024rethinking}  &  CVPR 2024 & 81.0 & 98.2 & 97.9 & 76.9 & 89.8 & 96.9 & 84.1 & 84.2 & 88.6 \\
FreqNet ~\cite{tan2024frequency} & AAAI 2024 & 89.6 & 98.8 & 98.6 & 66.8 & 86.5 & 97.3 & 75.8 & 81.4 & 86.8 \\
FatFormer~\cite{liu2024forgery} & CVPR 2024 & 92.7 & 100.0 & 99.9 & 75.9 & 88.0 & 99.9 & 98.8 & 55.8 & 88.9 \\
DRCT~\cite{chendrct} & ICML 2024 & 91.5&95.0&94.4&79.4&89.2&94.7&90.0&81.7&89.5 \\
\hline
\rowcolor{aliceblue} Ours & -- & 82.4 &  99.8 & 99.8 & 78.7 & 93.3 & 97.4 & 91.7 & 77.6 & 91.1 \\
\bottomrule
    \end{tabular}
}
  \label{tab:SOTA3}
\end{table*}

\subsection{Results on GenImage Benchmark} 

In our manuscript, we present the benchmarking outcomes of the UniversalFakeDetect Dataset. 
Additionally, we report the results obtained from another widely utilized benchmark known as GenImage \cite{zhu2024genimage}.
This GenImage dataset predominantly utilizes the Diffusion model for image generation, incorporating models such as Midjourney \cite{Midjourney}, SDv1.4 \cite{rombach2022high}, SDv1.5 \cite{rombach2022high}, ADM \cite{dhariwal2021diffusion}, GLIDE \cite{nichol2021glide}, Wukong \cite{Wukong}, VQDM \cite{gu2022vector}, and BigGAN \cite{BigGAN}. Following the settings defined for GenImage, we designate SDv1.4 as the training set and the remaining models as the test set.
Given the diverse image sizes within the GenImage dataset, images with a size smaller than 224 pixels are duplicated and subsequently cropped to 224 pixels, following \cite{tan2024rethinking}. We employ the same setting to re-implement FreqNet, FatFormer, and NPR, and also report the results of UnivFD and DRCT from \cite{chendrct}.

The results on the GenImage dataset are presented in Table \ref{tab:SOTA3}. When SDv1.4 is employed as the training set, our method attains an overall accuracy rate of {91.1\%} across the entire test set. Compared to similar methods that utilize CLIP as the backbone, such as UnivFD and FatFormer, our approach improves accuracy by {11.6\%} and {2.2\%}, respectively. Moreover, when contrasted with the latest state-of-the-art (SOTA) method DRCT (ICML 2024), the proposed method achieves a {1.6\%} enhancement in accuracy. This clearly indicates that our method demonstrates superior generalization capabilities and achieves SOTA performance on the GenImage benchmark.

\begin{table}
   
    \renewcommand\arraystretch{1.25}
    \centering
    \caption{
        \textbf{Cross-dataset generalization evaluations with existing SOTA video detectors}. The results of other detectors are directly cited from their original papers. The metric is video-level AUC.
    }
    \resizebox{1.0\columnwidth}{!}{
    \setlength\tabcolsep{5pt}
        \begin{tabular}{c|c|c|c|c} \toprule
        Methods & Venues & CDF-v2 & DFDC & Avg.\\ 
        \midrule
        LipForensics~\cite{haliassos2021lips} & CVPR 2021 & 0.824 & 0.735 & 0.780\\
        FTCN~\cite{zheng2021exploring} & ICCV 2021 & 0.869 & 0.740 & 0.805 \\
        HCIL~\cite{gu2022hierarchical} & ECCV 2022 & 0.790 & 0.692 & 0.741\\
        RealForensics~\cite{haliassos2022leveraging} & CVPR 2022 & 0.857 & 0.759 & 0.808 \\
        LTTD~\cite{guan2022delving} & NeurIPS 2022 & 0.893 & 0.804 & 0.849 \\
        AltFreezing~\cite{wang2023altfreezing} & CVPR 2023 & 0.895 & -- & -- \\
        TALL-Swin~\cite{xu2023tall} & ICCV 2023 & 0.908 & 0.768 & 0.838 \\
        StyleDFD~\cite{choi2024exploiting} & CVPR 2024 & 0.890 & -- & -- \\
        NACO~\cite{zhang2024learning} & ECCV 2024 & 0.895 & 0.767 & 0.831 \\
    
        \midrule
        
        Ours & -- & 0.956 & 0.843 & 0.900\\

        \bottomrule
      \end{tabular}
    }
    \label{tab:cmp_video_sota}
    \end{table}

\subsection{Comparison with Existing Video Detectors}
In the \textbf{manuscript}, we mainly compare our method with image detectors. Here, we provide an individual result to compare our approach with existing SOTA video detectors. 
Following \cite{zhang2024learning,xu2023tall}, we conduct evaluations on the widely-used CDF-v2~\cite{li2019celeb} and DFDC~\cite{dfdc} using the video-level AUC metric.
We have considered both the classical detectors such as LipForensics and the latest SOTA detectors such as NACO (ECCV'24) for a comprehensive comparison.
Results in Tab.~\ref{tab:cmp_video_sota} demonstrate that our image-based approach achieves higher generalization performance in both CDF-v2 and DFDC, improving {4.5\%} and {3.9\%} points than the second-best video-based models.
This further validates the superior generalization performance of our approach.

\begin{table}
\centering
\caption{\textbf{Additional results of cross-manipulation evaluation on FF++ (c23).} 
  Following \cite{miao2023f,luo2023forgery}, we conduct evaluations by training on the other three manipulated methods while testing on the remaining one. Specifically, GID-DF means training on the other three manipulated methods (FF-F2F, FF-FS, FF-NT) while testing on the FF-DF. 
  Results of other methods are cited from \cite{miao2023f,luo2023forgery}.}
\resizebox{1.0\columnwidth}{!}{
\setlength{\tabcolsep}{11pt}
\begin{tabular}{c|cc|cc}
\toprule
\multirow{2}{*}{Methods} & \multicolumn{2}{c|}{GID-DF} & \multicolumn{2}{c}{GID-F2F} \\ \cline{2-5} 
                        & Acc             & AUC             & Acc             & AUC                          \\ \hline
EfficientNet \cite{tan2019efficientnet}            & 82.40           & 91.11                            & 63.32           & 80.10 \\
MLGD \cite{li2018learning}                    & 84.21           & 91.82                    & 63.46           & 77.10                    \\
LTW \cite{sun2021domain}                     & 85.60           & 92.70                    & 65.60           & 80.20                    \\
DCL \cite{sun2022dual}                     & 87.70           & 94.90                    & 68.40           & 82.93                      \\
M2TR \cite{wang2022m2tr}                    & 81.07           & 94.91                 & 55.71           & 76.99                    \\
F3Net \cite{qian2020thinking}                  & 83.57           & 94.95                    & 61.07           & 81.20                  \\
F2Trans \cite{miao2023f}             & 89.64           & 97.47                    & 81.43           & 90.55                 \\ 
CFM \cite{luo2023beyond}      & 85.00           & 92.74                 & 76.07           & 84.55                   \\
FA-ViT \cite{luo2023forgery}                     & 92.86          & 98.10     & 82.57     & 91.20     \\ 
\hline
Ours                     & 95.71          & 99.26     & 85.71     & 93.83     \\ 
\bottomrule
\end{tabular}
}
\label{cross-mani}
\end{table}

\subsection{Robustness Evaluation}
\label{robust}

\begin{figure*}[!t] 
\centering 
\includegraphics[width=0.9\textwidth]{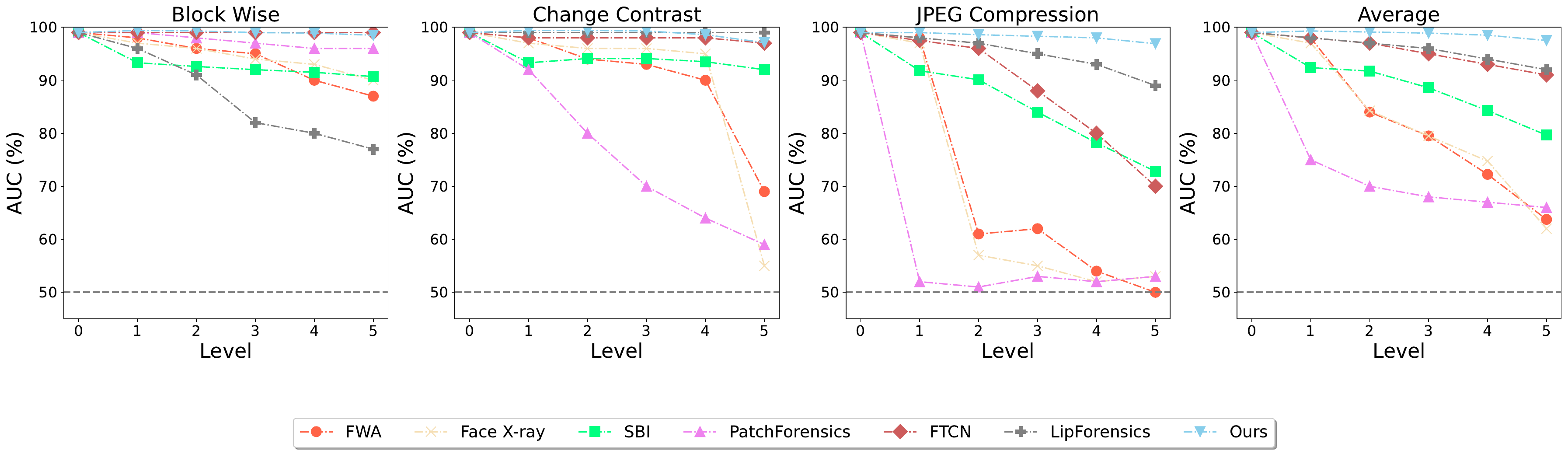} 
\caption{
\textbf{Robustness to unseen perturbations.} We present video-level AUC for five distinct degradation levels across three types of perturbations in~\cite{jiang2020deeperforensics}. 
} 
\label{fig:dongge} 
\vspace{-2mm}
\end{figure*}

To evaluate our model's robustness to random perturbations, we adopt the methodology used in previous studies~\cite{haliassos2021lips,zheng2021exploring}, which involves examining three distinct types of degradation: Block-wise distortion, Change contrast, and JPEG compression. We apply each of these perturbations at five different levels to assess the model's robustness under varying degrees of distortion, following~\cite{chen2022ost,yan2024transcending}. The video-level AUC results for these unseen perturbations, using the model trained on FF++ (c23), are depicted in Fig.~\ref{fig:dongge}. Generally, our approach shows higher results than other methods, demonstrating the better robustness of our approach than other models.

\begin{table}[t]
  \centering
  \caption{\textbf{Ablation studies on deepfake image detection (Cross-dataset) regarding different vision foundation models (VFMs) were used}. All models are trained on FF++ (c23) and tested on CDF-v2 and SimSwap.
  }
  \resizebox{1.0\columnwidth}{!}{
    \setlength\tabcolsep{5pt}
  \begin{tabular}{c|c|c|c|c|c} 
  \toprule
    VFMs & \#Params & \#ImgSize & CDF-v2 & SimSwap & Avg. \\ 
    \midrule
    BEIT-v2~\cite{peng2022beit} & 303M & 224 & 0.855 & 0.821 & 0.838 \\
    + Ours & 0.14M & 224 & 0.894 & 0.850 & 0.872 \\ 
    \midrule
    SigLIP~\cite{zhai2023sigmoid} & 316M & 256 & 0.877 & 0.713 & 0.795 \\
    + Ours & 0.19M & 256 & 0.895 & 0.778 & 0.867 \\ 
    \midrule
    CLIP~\cite{clip_paper} & 307M & 224 & 0.857 & 0.860 & 0.859 \\
    + Ours & 0.19M & 224 & 0.956 & 0.926 & 0.941 \\
    \bottomrule
  \end{tabular}
  }
  \label{tab:backbone_dfd}
  
\end{table}

\begin{table}[t]
  \centering
  \caption{\textbf{Ablation studies on synthetic image detection (UniversalFakeDetect Dataset) regarding different vision foundation models (VFMs) were used}. All models are trained on ProGAN's images and tested on 19 different generative models' images.
  }
  \resizebox{1.0\columnwidth}{!}{
    \setlength\tabcolsep{10pt}
  \begin{tabular}{c|c|c|c|c} 
  \toprule
    VFMs & \#Params & \#ImgSize & mAP & mAcc \\ 
    \midrule
    BEIT-v2~\cite{peng2022beit} & 303M & 224 & 93.50 & 79.11  \\
    + Ours & 0.14M & 224 & 97.39 & 83.66 \\ 
    \midrule
    SigLIP~\cite{zhai2023sigmoid} & 316M & 256 & 94.30 & 81.23 \\
    + Ours & 0.19M & 256 & 96.24 & 90.46 \\ 
    \midrule
    CLIP~\cite{clip_paper} & 307M & 224 & 97.95 & 86.22  \\
    + Ours & 0.19M & 224 & 99.41 & 95.19 \\
    \bottomrule
  \end{tabular}
  }
  \label{tab:backbone_aigc}
\end{table}

\section{Additional cross-method evaluation}
The capability of face forgery detectors to generalize to new manipulation methods is crucial in practical, real-world applications.
In our manuscript, we present cross-method evaluations using the DF40 dataset \cite{yan2024df40}. Specifically, we train the models with four manipulation methods from FF++ (c23) and then test them on the other eight manipulation techniques provided in DF40.
Furthermore, we conduct an additional cross-method evaluation following the protocol introduced in \cite{sun2022dual, miao2023f, luo2023forgery}. This protocol involves training the model on diverse manipulation types of samples and subsequently testing it on unknown manipulation methods. The results of this evaluation are reported in Tab. \ref{cross-mani}. It is evident that our proposed method attains remarkable performance in cross-manipulation evaluation. In terms of accuracy (ACC), it outperforms the latest SOTA detector FA-ViT by 2.85\% on GID-DF and 3.14\% on GID-F2F, respectively.

\begin{table}[t]
  \centering
  \caption{\textbf{Ablation studies on deepfake image detection (Cross-dataset) regarding different ViT architectures were used}. We employ the two architectures implemented in the original paper of CLIP \cite{clip_paper} for experiments.
  All models are trained on FF++ (c23).
  }
  \resizebox{1.0\columnwidth}{!}{
  \setlength\tabcolsep{5pt}
  \begin{tabular}{c|c|c|c|c|c} 
  \toprule
    VFMs & \#Params & \#ImgSize & CDF-v2 & SimSwap & Avg. \\ 
    \midrule
    CLIP-Base/16 & 86M & 224 & 0.854 & 0.833 & 0.844 \\
    + Ours & 0.07M & 224 & 0.915 & 0.919 & 0.917 \\
    \midrule
    CLIP-Large/14 & 307M & 224 & 0.857 & 0.860 & 0.859 \\
    + Ours & 0.19M & 224 & 0.956 & 0.926 & 0.941 \\
    \bottomrule
  \end{tabular}
  }
  \label{tab:scale_dfd}
  
\end{table}

\begin{table}[t]
  \centering
  \caption{\textbf{Ablation studies on synthetic image detection (UniversalFakeDetect Dataset) regarding different architectures were used}. All models are trained on ProGAN's images and tested on 19 different kinds of generative models' images.
  }
  \resizebox{1.0\columnwidth}{!}{
    \setlength\tabcolsep{10pt}
  
  \begin{tabular}{c|c|c|c|c} 
  \toprule
    VFMs & \#Params & \#ImgSize & mAP & mAcc \\ 
    \midrule
    CLIP-Base/16 & 86M & 224 & 96.25 & 82.52 \\
    + Ours & 0.07M & 224 & 98.47 & 88.46 \\ 
    \midrule
    CLIP-Large/14 & 307M & 224 & 97.95 & 86.22 \\
    + Ours & 0.19M & 224 & 99.41 & 95.19 \\
    \bottomrule
  \end{tabular}
  }
  \label{tab:scale_aigc}
  
\end{table}

\begin{figure*}[!ht] 
\centering 
\includegraphics[width=1.0\textwidth]{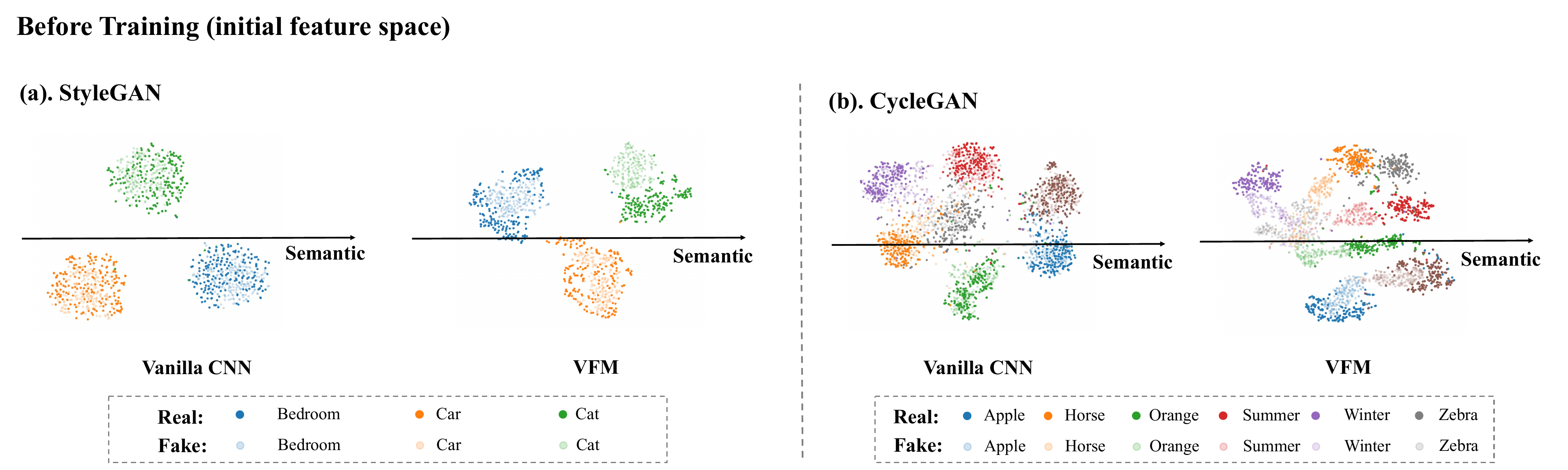} 
\vspace{-3mm}
\caption{
\textbf{t-SNE visualizations of the \textit{initial} latent feature spaces between vanilla CNN~\cite{wang2020cnn} and CLIP~\cite{clip_paper}}. We show that both the pre-trained CNN and CLIP can identify different semantic objects.
} 
\label{fig:init_tsne} 
\vspace{-3mm}
\end{figure*}

\begin{figure*}[!ht] 
\centering 
\includegraphics[width=1.0\textwidth]{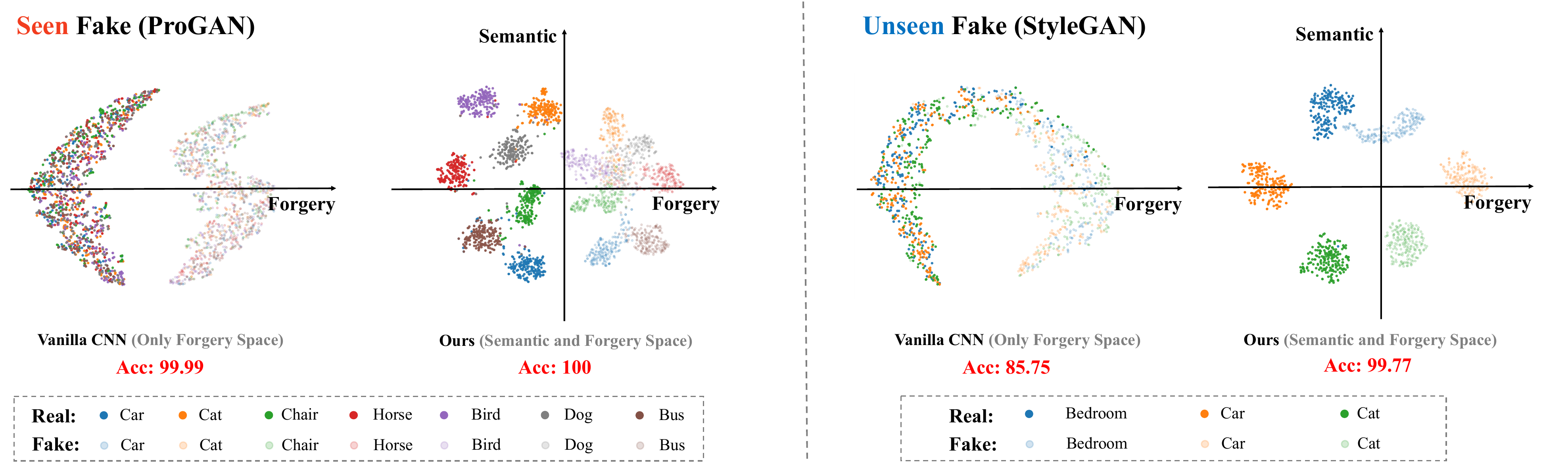} 
\vspace{-3mm}
\caption{
\textbf{t-SNE visualizations of the latent feature spaces between vanilla CNN~\cite{wang2020cnn} and ours.} We use the testing set of ProGAN and StyleGAN within UniversalFakeDetect Dataset~\cite{wang2020cnn} for visualization. We see that after fine-tuning on AIGI data, the baseline quickly shortcuts to the fake and ``forgets" the pre-existing semantic knowledge, thereby resulting in a highly constrained feature space.
} 
\label{fig:tsne_progan_stylegan} 
\vspace{-3mm}
\end{figure*}


\subsection{Additional Ablation Studies}
\paragraph{Impact of Different Vision Foundation Models} 
We initialize the ViT backbone with several widely used pre-trained weights from different vision foundation models, including BEIT-v2 \cite{peng2022beit}, CLIP \cite{clip_paper}, and SigLIP~\cite{zhai2023sigmoid}. 
The results are shown in Tab.~\ref{tab:backbone_dfd} and Tab.~\ref{tab:backbone_aigc}. 
It is evident that our proposed approach improves the generalization performance of different pre-trained ViTs. On the other hand, we note that different initialization significantly impacts generalization performance, indicating the importance of choosing a suitable pre-trained initialization. Through empirical results, we discover that the ViT pre-trained on CLIP exhibits the highest performance in both deepfake detection and synthetic image detection tasks. Therefore, we choose CLIP as the default setting for our approach.

\paragraph{Impact of Different ViT Backbones} 
Here, we investigate the effects of different ViT architectures. Specifically, we consider two backbones that were implemented in the original paper of CLIP: ViT-Base-16 and ViT-Large-14.
We conduct evaluations on both deepfake detection and synthetic image detection benchmarks, as shown in Tab.~\ref{tab:scale_dfd} and Tab.~\ref{tab:scale_aigc}. Compared to fully fine-tuning the CLIP model, our proposed approach consistently demonstrates substantial enhancements in generalization performance across these ViT backbones. It is worth noting that CLIP-Large performs better than CLIP-Base by a notable margin. Based on this ablation experiment, we ultimately choose ViT-Large as our default backbone.

\section{Additional Analysis and Visualizations} \label{section3}

\subsection{Additional t-SNE Visualizations}
We further visualize the t-SNE of the seen fake ProGAN, unseen fake StyleGAN and useen CycleGAN for the comparison of vanilla CNN (Res-50~\cite{wang2020cnn}) and ours (see Fig.~\ref{fig:init_tsne}, Fig.~\ref{fig:tsne_progan_stylegan}).
As we can see from both two figures, our approach maximizes and preserves the pre-trained knowledge while fitting the forgery patterns during training, whereas the vanilla CNN overfits the seen fake method, learning forgery patterns only, thereby resulting in a highly low-ranked feature space (see Fig.~\ref{fig:pca_results} and Fig.~\ref{fig:real_fake_logits} of the manuscript for details) and causing the overfitting to seen forgery patterns in the training set.
Additionally, we see that the logit distribution of the vanilla CNN has a larger overlapping between fake and real, while ours is highly smaller, suggesting that our approach achieves a better generalization performance.

\subsection{Self-attention Map Visualizations}
\begin{figure*}[!ht] 
\centering 
\includegraphics[width=1.0\textwidth]{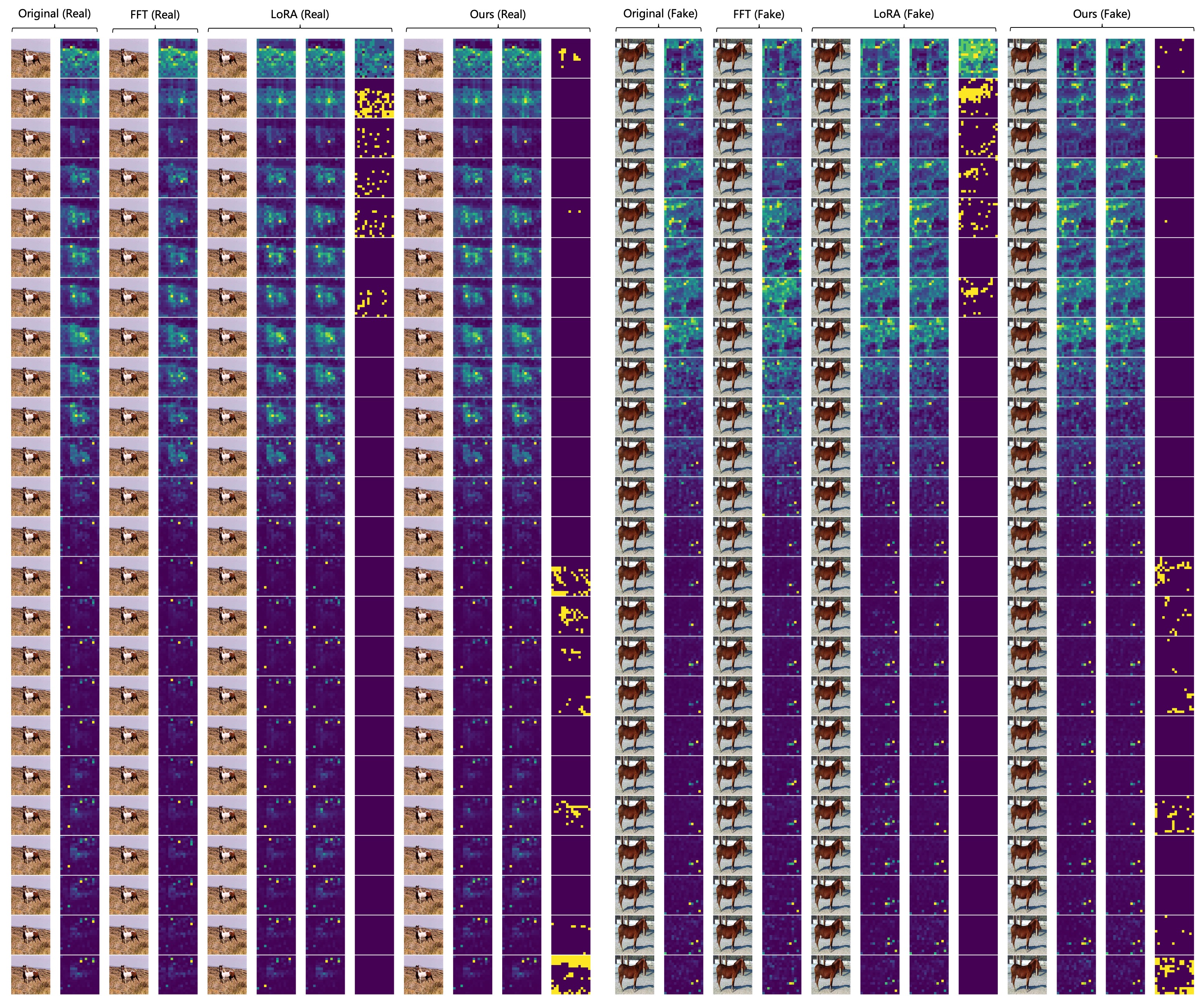} 
\vspace{-3mm}
\caption{\textbf{Self-attention map visualizations of UniversalFakeDetect Dataset~\cite{wang2020cnn}.} We visualize the fake image of CycleGAN part and the corresponding real image for each block of the CLIP-ViT model (there are a total of 24 blocks, with IDs gradually increasing from top to bottom).}
\label{fig:attnmap1} 
\vspace{-3mm}
\end{figure*}

\begin{figure*}[!ht] 
\centering 
\includegraphics[width=1.0\textwidth]{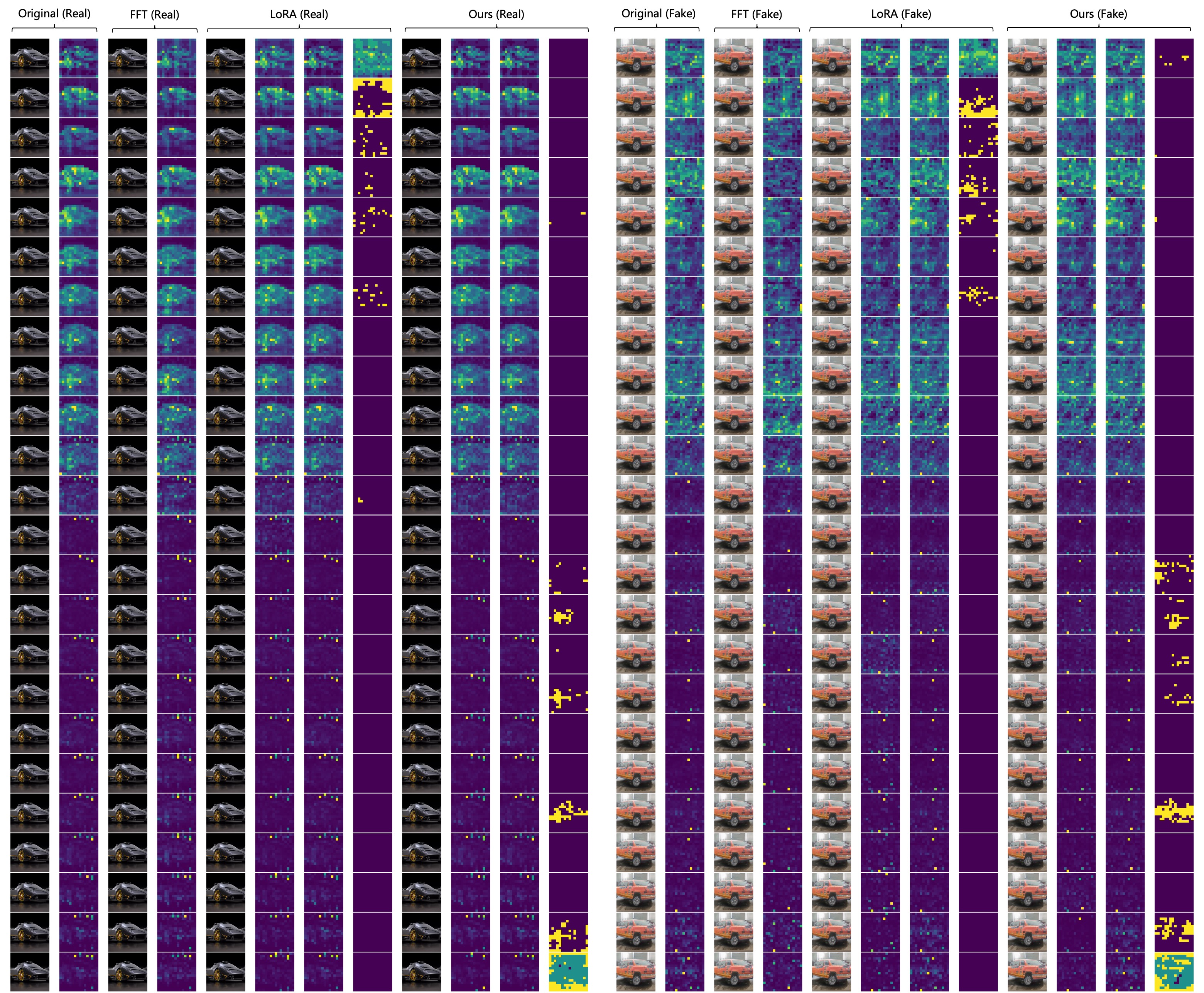} 
\vspace{-3mm}
\caption{\textbf{Self-attention map visualizations of UniversalFakeDetect Dataset~\cite{wang2020cnn}.} We visualize the fake image of ProGAN part and the corresponding real image for each block of the CLIP-ViT model (there are a total of 24 blocks, with IDs gradually increasing from top to bottom).}
\label{fig:attnmap2} 
\vspace{-3mm}
\end{figure*}

\begin{figure*}[!ht] 
\centering 
\includegraphics[width=1.0\textwidth]{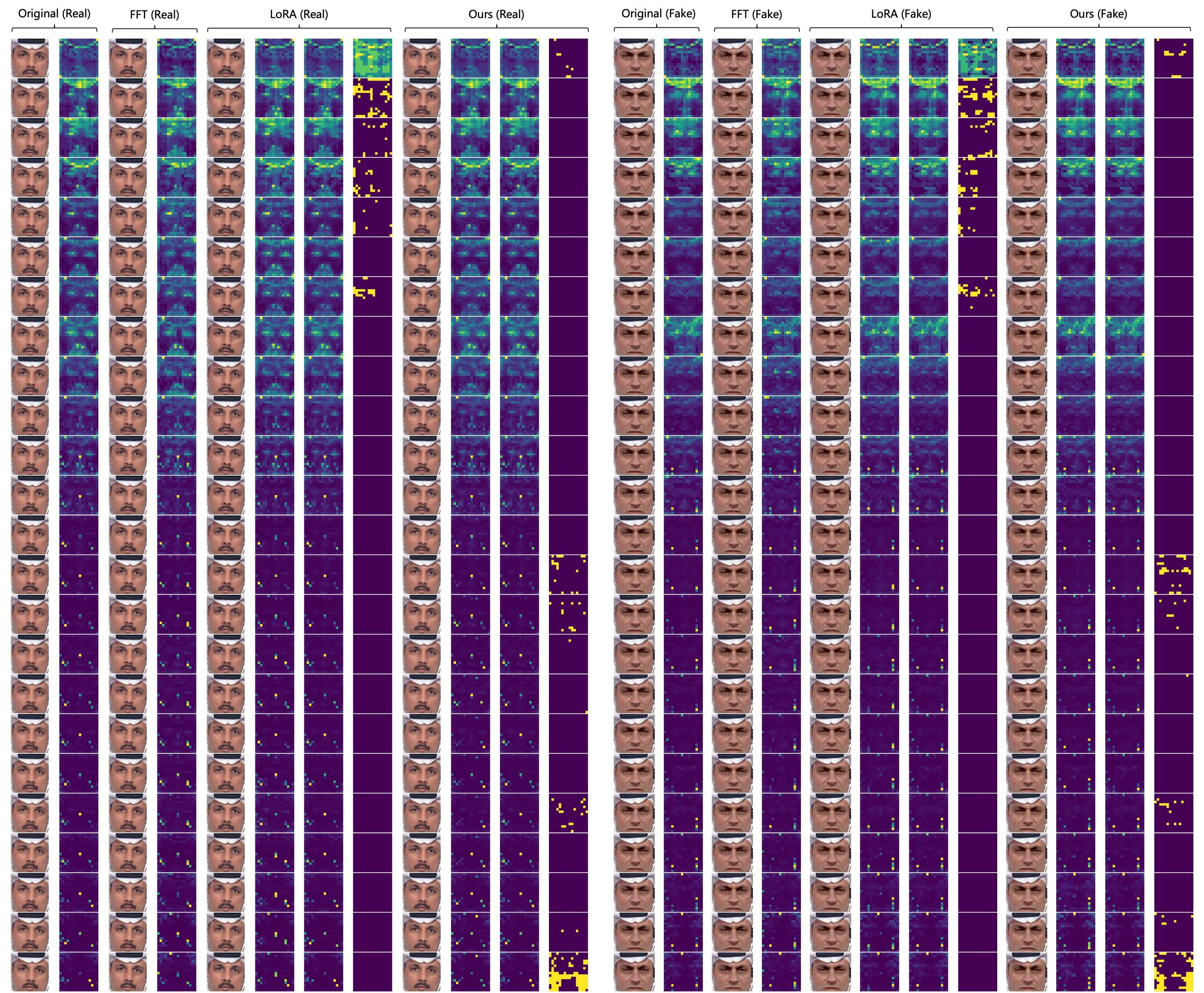} 
\vspace{-3mm}
\caption{\textbf{Self-attention map visualizations of UniversalFakeDetect Dataset~\cite{wang2020cnn}.} We visualize the fake image of DeepFake part and the corresponding real image for each block of the CLIP-ViT model (there are a total of 24 blocks, with IDs gradually increasing from top to bottom).}
\label{fig:attnmap3} 
\vspace{-3mm}
\end{figure*}

\begin{figure*}[!ht] 
\centering 
\includegraphics[width=1.0\textwidth]{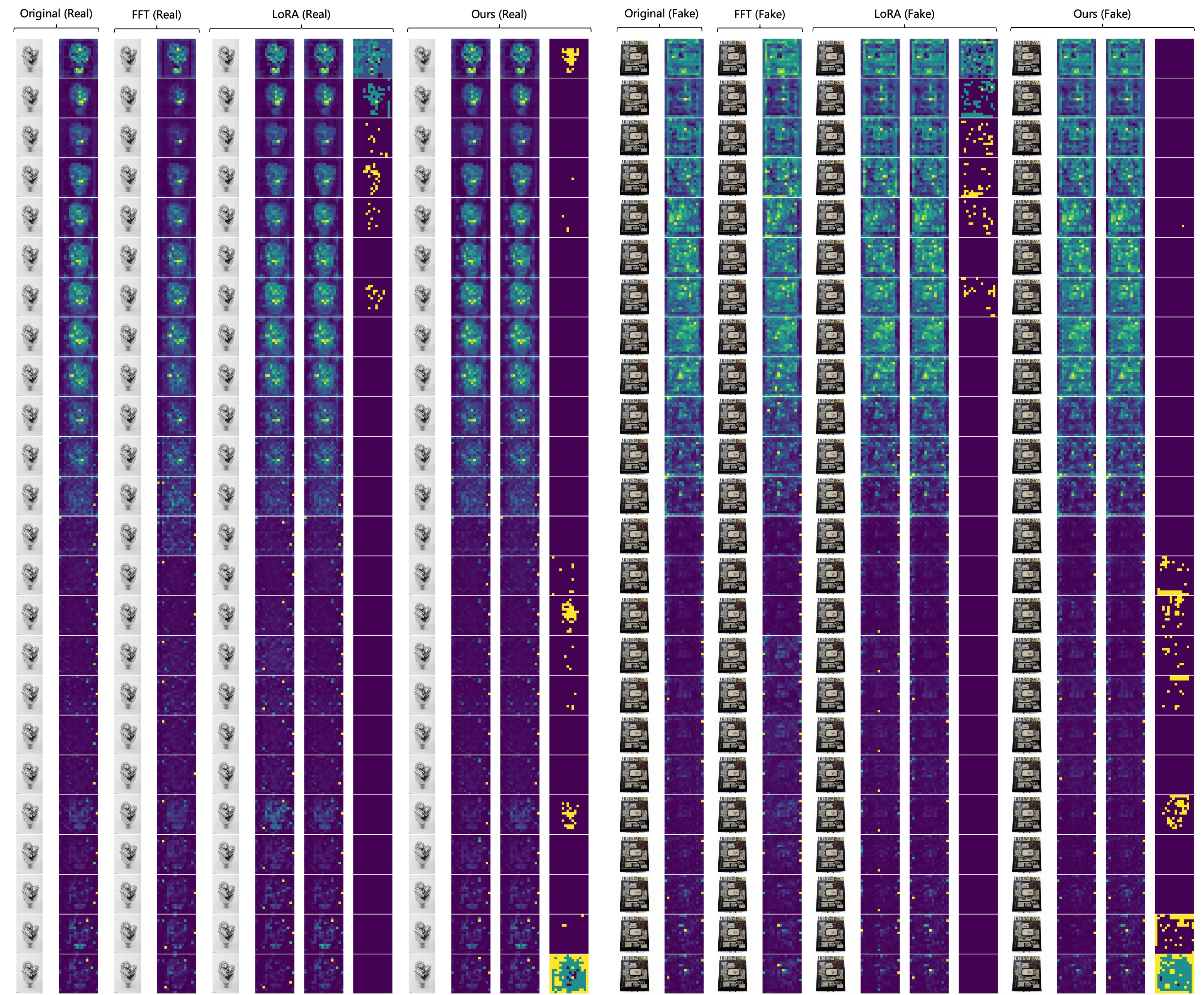} 
\vspace{-3mm}
\caption{\textbf{Self-attention map visualizations of UniversalFakeDetect Dataset~\cite{wang2020cnn}.}We visualize the fake image of LDM part and the corresponding real image for each block of the CLIP-ViT model (there are a total of 24 blocks, with IDs gradually increasing from top to bottom).}
\label{fig:attnmap4} 
\vspace{-3mm}
\end{figure*}

Here, we perform the self-attention maps visualization of the original CLIP-ViT model (Original), the fully fine-tuned CLIP-ViT model (FFT), the LoRA-trained CLIP-ViT model (LoRA), and our proposed orthogonal trained CLIP-ViT model (Ours) on the UniversalFakeDetect dataset (see Fig.~\ref{fig:attnmap1}, Fig.~\ref{fig:attnmap2}, Fig.~\ref{fig:attnmap3} and Fig.~\ref{fig:attnmap4}). Specifically, for each block of the ViT, the self-attention map denotes the self-attention coefficient matrix calculated between the [CLS] token and the patch tokens. In the case of the LoRA component, the self-attention maps are generated from left to right using the original + LoRA weights, the original weights, and the LoRA weights, respectively. In the case of the Ours component, the self-attention maps are generated from left to right using the principal + residual weights, the principal weights, and the residual weights, respectively. Surprisingly, we observe that the semantic information is primarily concentrated in the earlier blocks, and our proposed approach establishes orthogonality between the semantic subspace and the learned forgery subspace at the level of the self-attention map. It further explains that our proposed approach can better preserve the pre-trained knowledge while learning fake patterns.

\section{Limitation and Future Work} \label{section4}
\vspace{-2mm}
The core idea of this paper is to decompose the original feature space into two orthogonal subspaces for preserving pre-trained knowledge while learning the forgery.
In our manuscript and supplementary, we have conducted extensive experiments and in-depth analysis on both deepfake and synthetic image detection benchmarks, showing the superior advantages in both generalization and efficiency.
One limitation of our work is that our approach regards all fake methods in one class during training real/fake classifiers, potentially ignoring the specificity and generality of different fake methods.

In the future, we plan to expand our approach into a \textit{incremental learning} framework, where each fake method will be regarded as ``one SVD branch", ensuring the orthogonality between different fake methods, thereby avoiding the severe forgetting of previous learned fake methods.
This extension design will help our approach better address the future deepfake types in the real-world scenario.
Additionally, although our work's scope mainly focuses on deepfake and synthetic image detection, our approach also has the potential to be applied to other similar fields such as face anti-spoofing, anomaly detection, etc.
Furthermore, we hope our proposed approach can inspire future research in developing better orthogonal modeling strategies.

\textbf{Ethics \& Reproducibility.} 
All of the facial images that are utilized are sourced from publicly available datasets and are accompanied by appropriate citations. This guarantees that there is no infringement upon personal privacy. We will make all codes and checkpoints available for public access upon acceptance.

\bibliographystyle{icml2025}

\end{document}